\documentclass[10pt]{article}

\usepackage[square,numbers]{natbib} 

\usepackage{microtype}
\usepackage{graphicx}
\usepackage{booktabs} 
\usepackage{multirow}
\usepackage{amsmath}
\usepackage{amsfonts}
\usepackage{tabularx}
\usepackage{amssymb}
\usepackage{xcolor}
\usepackage{dsfont}
\newcommand{\un}{\mathds{1}}
\usepackage{algorithm}
\usepackage{subcaption}
\usepackage{natbib}
\usepackage{nicefrac}

\usepackage{algorithmicx}
\usepackage{algpseudocode}
\usepackage{graphics}
\usepackage{multicol}

\newcommand{\cccmt}[1]{{\textcolor{brown}{[#1]}}}

\newcommand{\HTalgo}{{\fontfamily{qcr}\small\textbf{LHTR}}}
\newcommand{\NN}{{\fontfamily{qcr}\small\textbf{NN model}}}
\newcommand{\geneliex}{{\fontfamily{qcr}\small\textbf{GENELIEX}}}

\algnewcommand\algorithmicinput{\textbf{INPUT:}}
\algnewcommand\INPUT{\item[\algorithmicinput]}

\algnewcommand\algorithmicoutput{\textbf{OUTPUT:}}
\algnewcommand\OUTPUT{\item[\algorithmicoutput]}
\usepackage{changes}

\algnewcommand\algorithminit{\textbf{Initialization:}}
\algnewcommand\Initialize{\item[\algorithminit]}

\algnewcommand\algorithStandardization{\textbf{Threshold:}}
\algnewcommand\Standardization{\item[\algorithStandardization]}

\algnewcommand\algorithoptim{\textbf{Optimization:}}
\algnewcommand\Optimization{\item[\algorithoptim]}

\usepackage{tikz}
\usetikzlibrary{patterns}
\usepackage{hyperref}
\usepackage{url}
\usepackage{wrapfig,lipsum,booktabs}
\usepackage{microtype}
\usepackage{graphicx}
\usepackage{microtype}

\newcommand{\Prob}[1]{\mathbb{P}\left\{ #1 \right\} }
\newcommand{\PP}{\Prob}
\newcommand{\given}[1][{}]{\;\middle\vert\;{#1} }

\newcommand{\RR}{\mathbb{R}}
\newcommand{\rset}{\RR}
\newtheorem{theorem}{Theorem}
\newtheorem{rema}{Remark}
\newcommand{\defeq}{\overset{\text{\tiny def}}{=}}
\newenvironment{remark}{\begin{rema} \rm}{\end{rema}}

\usepackage{soul}

\usepackage{color}
\usepackage{changes}
\definechangesauthor[color=purple]{Pierre}
\definechangesauthor[color=blue]{Hamid}
\definechangesauthor[color=orange]{Anne}
\definechangesauthor[color=red]{Chloe}

\newcommand\addeda[1]{\added[id=Anne]{#1}}

\usepackage[final]{neurips_2020}
\hypersetup{colorlinks,linkcolor={blue},citecolor={blue},urlcolor={blue}}  

\title{Heavy-tailed Representations, Text Polarity Classification \& Data Augmentation}
\author{ Hamid Jalalzai\thanks{Both authors contributed equally} \\
	LTCI, T\'el\'ecom Paris\\
	Institut Polytechnique de Paris\\
	\texttt{hamid.jalalzai@telecom-paris.fr} \\
	\And
	Pierre Colombo\footnotemark[1] \\
	IBM France \\
	LTCI, T\'el\'ecom Paris \\
	Institut Polytechnique de Paris\\
	\texttt{pierre.colombo@telecom-paris.fr} \\
	\And
	Chlo\'e Clavel \\
	LTCI, T\'el\'ecom Paris\\
	Institut Polytechnique de Paris\\
	\texttt{chloe.clavel@telecom-paris.fr} \\
	\And
	Eric Gaussier \\
	Univ. Grenoble Alpes, CNRS, Grenoble INP, LIG\\
	\texttt{eric.gaussier@imag.fr} \\
	\And
	Giovanna Varni \\
	LTCI, T\'el\'ecom Paris\\
	Institut Polytechnique de Paris\\
	\texttt{giovanna.varni@telecom-paris.fr} \\
	\And
	Emmanuel Vignon \\
	IBM France\\
	\texttt{emmanuel.vignon@fr.ibm.com} \\
	\And
	Anne Sabourin \\
	LTCI, T\'el\'ecom Paris\\
	Institut Polytechnique de Paris\\
	\texttt{anne.sabourin@telecom-paris.fr} \\
}
\date{}

\begin{document}
\maketitle

\begin{abstract}
The dominant approaches to text representation in natural language rely on learning embeddings on massive corpora which have convenient properties such as compositionality and distance preservation. In this paper, we develop a novel method to learn a heavy-tailed embedding with desirable regularity properties regarding the distributional tails, which allows to analyze the points far away from the distribution bulk using the framework of multivariate extreme value theory. In particular, a classifier dedicated to the tails of the proposed embedding is obtained which exhibits a \textit{scale invariance} property exploited in a novel text generation method for label preserving dataset augmentation. Experiments on synthetic and real text data show the relevance of the proposed framework and confirm that this method generates meaningful sentences with controllable attributes, \textit{e.g.} positive or negative sentiments.
\end{abstract}

\section{Introduction}
\label{sec:intro}
\begin{figure}[t]
    \centering
    \includegraphics[width=0.20\textwidth]{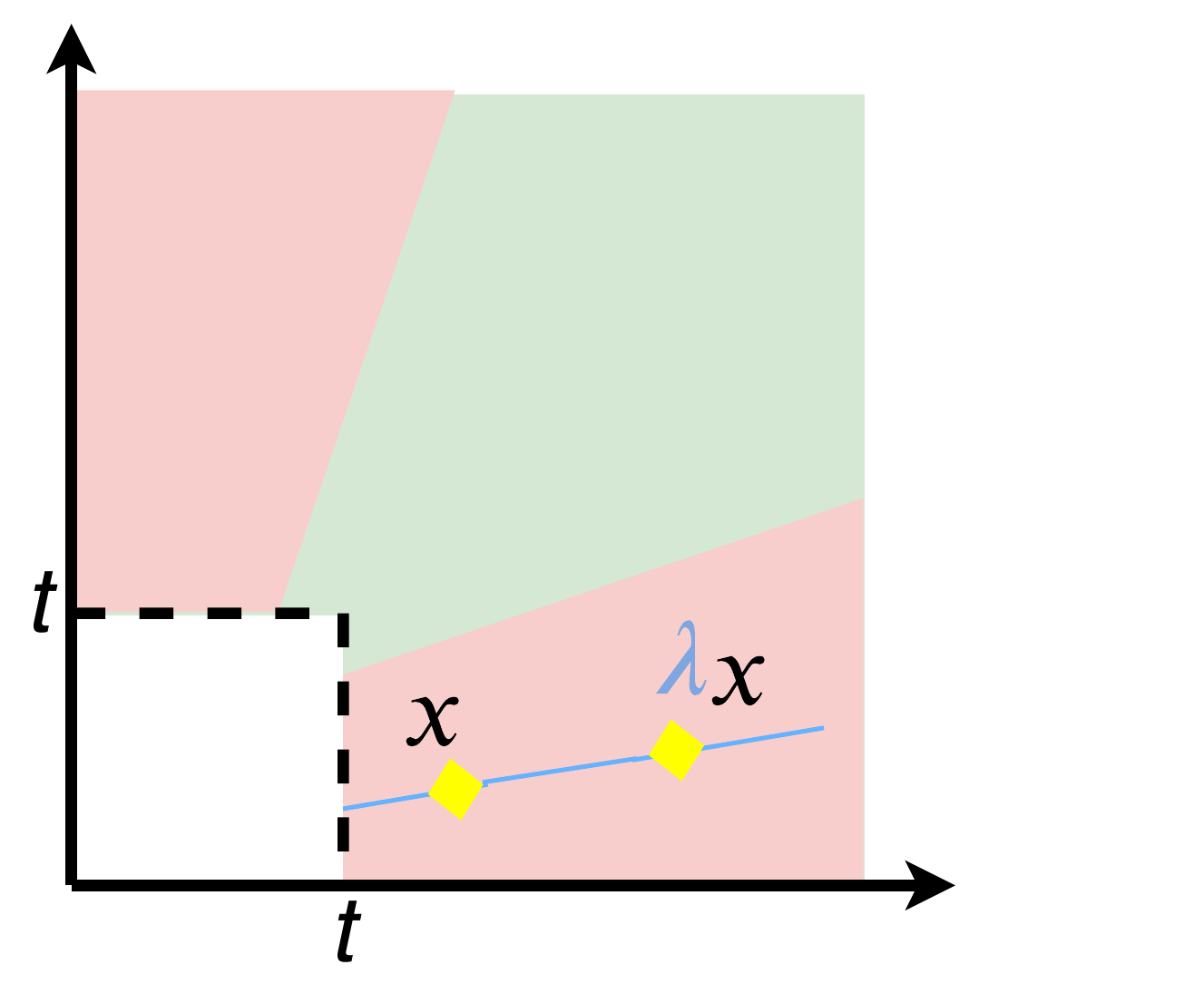} 
    \caption{Illustration of angular classifier $g$  dedicated to extremes $\{x, \|x\|_\infty \geq t \}$ in $\rset_+^2$. The red and green truncated cones are respectively labeled as $+1$ and $-1$ by $g$.}  
    \label{fig:my_label}
\end{figure}

Representing the meaning of natural language in a mathematically grounded way is a scientific challenge that has received increasing attention with the explosion of digital content and text data in the last decade. Relying on the richness of contents, several embeddings have been proposed  \cite{elmo,gpt,devlin2018bert} with  demonstrated efficiency for the considered tasks when learnt on massive datasets. However, none of these embeddings take into account the fact that word frequency distributions  are heavy tailed \cite{extrem_2,extrem_3, mandelbrot1953informational}, so that extremes are naturally present in texts (see also Fig.~\ref{fig:pearson_bert_yelp} and \ref{fig:pearson_bert_amazon} in the supplementary material). Similarly, \cite{extrem_1} shows that, contrary to image taxonomies, the underlying distributions for words and documents in large scale textual taxonomies are also heavy tailed. Exploiting this information, several studies, as \cite{clinchant2010information, madsen2005modeling}, were able to improve text mining applications by accurately modeling the tails of textual elements. \\
In this work, we rely on the framework of multivariate extreme value analysis, based on extreme value theory (EVT) which focuses on the distributional tails.
EVT is valid under a regularity assumption  which amounts to a homogeneity property above large thresholds: the tail behavior of the considered  variables must be well approximated by a power law, see Section~\ref{sec:MEVT} for a rigorous statement. The tail region (where  samples are considered as extreme) of the input  variable $x\in\rset^d$ is of the kind $\{\|x \| \geq t\}$,  for a large threshold $t$.   
The latter is typically chosen such that a small but non negligible proportion of the data is considered as extreme, namely $25\%$ in our experiments.
A major advantage of this framework in the case of labeled data \cite{jalalzai2018binary} is that  classification on the tail regions
may be performed using the  angle $\Theta(x) = \|x\|^{-1} x$ only, see Figure~\ref{fig:my_label}.   The main idea behind the present paper is to take advantage of the scale invariance for two tasks regarding sentiment analysis of text data: \emph{(i)}~Improved classification of extreme inputs,  
\emph{(ii)}~Label preserving data augmentation, as the most probable label of an input $x$ is unchanged by multiplying $x$ by $\lambda>1$. 

\vspace{-0.1cm}
EVT in a machine learning framework has received increasing attention in the past few years. Learning  tasks considered so far include  anomaly detection \cite{Roberts99,Roberts2000,Clifton2011,goix2016sparse,thomas2017anomaly},  anomaly clustering~\cite{chiapino2019multivariate}, unsupervised learning~\cite{goix2015learning}, online learning~\cite{carpentier2014extreme,achab2017max}, dimension reduction  and support identification~\cite{goix2017sparse,chiapino2016feature,chiapino2019identifying,mexico}. 
The present paper builds upon the methodological framework proposed by  \citet{jalalzai2018binary} for classification in extreme regions.  
The goal of~\citet{jalalzai2018binary} is to improve the performance of classifiers  $\widehat g(x)$ issued from Empirical Risk Minimization (ERM) on the tail regions $\{ \| x\| >t\}$ 
Indeed, they argue that  for very large  $t$, there is no guarantee that 
$\widehat g$ 
would perform well conditionally to $\{ \|X\|>t \}$,  precisely because of the scarcity of such examples in the training set.  They thus propose to train a specific classifier dedicated to extremes leveraging the probabilistic structure of the tails. 
\citet{jalalzai2018binary} demonstrate the usefulness of their framework with simulated and some real world 
datasets. However, there is no reason to assume that the previously mentioned text embeddings satisfy the required regularity assumptions. 
The aim of the present work is to extend \cite{jalalzai2018binary}'s methodology to datasets which do not satisfy their assumptions, in particular to text datasets embedded by state of the art techniques. This is achieved by the algorithm \emph{Learning a Heavy Tailed Representation} (in short {\fontfamily{qcr}\small\textbf{LHTR}}) which learns a transformation mapping the input data $X$ onto a random vector $Z$ which does satisfy the aforementioned  assumptions.
The transformation is learnt by an adversarial strategy \cite{Goodfellow-et-al-2016}. 

 In Appendix~\ref{explainabilite} we propose an interpretation of the extreme nature of an input in both  \HTalgo\ and  BERT representations. In a word, these sequences are longer and are more difficult to handle (for next token prediction and classification tasks) than non extreme ones.

 Our second contribution is  a novel data augmentation mechanism \geneliex\   which takes advantage of the scale invariance properties of $Z$ to generate synthetic sequences 
 that keep invariant the attribute of the original sequence. 
Label preserving data augmentation 
is an effective solution to the data scarcity problem and is an efficient pre-processing step for moderate 
dimensional datasets 
\cite{computer_vision_1,eda}. Adapting these methods to  NLP problems remains a challenging  issue. 
The problem  consists in constructing a transformation $h$ such that for any sample $x$ with label $y(x)$, the generated sample $h(x)$ would remain label consistent:
$~y\big(h(x)\big) = y(x) $  \cite{ratner2017learning}.  
The dominant approaches for text data augmentation rely on word level transformations such as 
synonym replacement, slot filling, swap deletion \cite{eda} using external resources such as wordnet \cite{miller1995wordnet}. 
Linguistic based approaches can also be combined with vectorial representations provided by language models \cite{naacl_baseline}. However, to the best of our knowledge, building a vectorial transformation without using any external linguistic resources remains an open problem.
%
In this work, as the label $y\big(h(x)\big)$ 
is unknown as soon as $h(x)$ does not belong to the training set,  we address this issue by learning  
both an embedding $\varphi$  and a classifier $g$ satisfying a relaxed version of the problem above mentioned, namely $\forall \lambda \ge 1$
\begin{equation}    \label{eq:invariance_dilation}
    g\big(h_\lambda(\varphi(x))\big) = g\big(\varphi(x)\big). 
\end{equation}
 For mathematical reasons which will appear clearly in Section~\ref{sec:classifExtremes}, $h_\lambda$ is chosen as the  homothety with scale factor $\lambda$, $h_\lambda (x) =\lambda x$.  
 In this paper, we work with output vectors issued by BERT 
 \cite{devlin2018bert}.
 BERT and its variants are currently the most widely used language model but we emphasize 
 that the proposed methodology could equally  be applied using any  other representation as input.
 BERT embedding does not satisfy the regularity properties required by EVT (see the results from statistical tests performed in  Appendix~\ref{BERT_notRV}) 
 Besides, there is no reason why a classifier $g$ trained on such embedding would be scale invariant, \emph{i.e.}  would satisfy  for a given sequence 
 $u$, embedded as $x$, $g(h_\lambda(x)) = g(x)$ $\forall \lambda \geq 1$.
 On the classification task, we demonstrate on two datasets of sentiment analysis that the embedding learnt by {\fontfamily{qcr}\small\textbf{LHTR}} on top of BERT is indeed following a heavy-tailed distribution. Besides, a classifier trained on the embedding learnt by {\fontfamily{qcr}\small\textbf{LHTR}} outperforms 
 the same classifier trained on BERT. On the dataset augmentation task, quantitative and qualitative experiments demonstrate the ability of \geneliex\  to generate new sequences 
 while preserving labels.

 The rest of this paper is organized as follows. Section~\ref{sec:MEVT} introduces the necessary background in multivariate extremes. The methodology we propose is detailed at length in Section~\ref{sec:Contribution}. Illustrative numerical experiments on both synthetic and real data are gathered in sections~\ref{sec:Exp}
 and~\ref{sec:Exp_g}. 
 Further comments and experimental results are provided in the supplementary material. 


\section{Background}
\label{sec:MEVT}
\subsection{Extreme values, heavy tails and  regular variation}\label{sec:extremes}

Extreme value analysis is a branch of statistics whose main focus is on 
 events characterized by an unusually high value of a monitored quantity.  A convenient  working assumption in EVT  is \emph{regular variation}. 
A real-valued random variable  $X$ 
 is regularly varying with index $\alpha>0$, a property denoted as  $RV(\alpha)$, 
if and only if there exists a function $b(t)>0$, with $b(t)\to \infty$ as $t\to \infty$,  such that  for any fixed $x>0$:
$
t\mathbb{P}\left\{X/b(t) > x  
\right\} \xrightarrow[t\rightarrow\infty]{}
x^{-\alpha} \,.
$
In the multivariate case $X = (X_1, \ldots, X_d ) \in \rset^d$, it is usually assumed that a preliminary component-wise  transformation has been applied so that each margin $X_j$ is $RV(1)$ with $b(t) = t$ and takes only positive values. $X$ is 
\emph{standard multivariate regularly varying} if  there exists a positive Radon measure $\mu$ on 
$[0,\, \infty]^d\backslash\{0\}$ 
\begin{equation}
\label{RV}
t \mathbb{P}\left\{ t^{-1} X \in  A\right\} \xrightarrow[t\rightarrow\infty]{}
\mu(A) , 
\end{equation}
for any Borelian set $A\subset [0,\infty]^d$ which is bounded away from $0$ and such that the limit measure $\mu$ of the boundary  $\partial A$ is zero.
For a complete introduction to the theory of Regular Variation, the reader may refer to \cite{resnick2013extreme}. 
The measure $\mu$ may  be understood as the limit distribution  of tail events. 
In~(\ref{RV}), $\mu$ is homogeneous of order $-1$, that is $\mu(tA) = t^{-1} \mu(A)$, $t>0, A\subset [0,\infty]^d\setminus\{0\}$. This scale invariance is key for our purposes, as detailed in  Section~\ref{sec:classifExtremes}.  
The main idea behind extreme value analysis is to learn relevant features of $\mu$ 
using the largest  available data.   
\subsection{Classification in extreme regions}\label{sec:classifExtremes} 
We now recall the classification setup for extremes as introduced in \cite{jalalzai2018binary}. Let  $(X, Y) \in \rset^d_{+} \times \{-1, 1\} $ be a random pair. 
Authors of \cite{jalalzai2018binary} 
assume standard regular variation for both classes,  
that is 
$ ~t \PP{X \in tA \given Y = \pm 1} \to \mu_{\pm}(A)$, where $A$ is as in~(\ref{RV}). 
Let $\|\,\cdot\,\|$ be any norm on $\rset^d$ and consider the risk of a classifier $g: \rset_+^d \to \{\pm1\}$ above a radial threshold $t$,
\begin{equation}
  \label{eq:extremeRisk}
L_t(g)= \PP{Y \neq g(X) \given \|X\|>t}.   
\end{equation}
The goal is to minimize the asymptotic risk in the extremes  $L_\infty(g) = \limsup_{t\to\infty} L_t(g)$. Using the scale invariance property of $\mu$,  under additional mild regularity assumptions concerning the regression function, namely uniform convergence to the limit at infinity, one can prove the following result (see \cite{jalalzai2018binary}, Theorem~1):  
  there exists a classifier $g_\infty^\star$ depending  on the pseudo-angle   $\Theta(x) = \|x\|^{-1} x $ only, that is $g_\infty^\star (x) = g_\infty^\star \big(\Theta(x)\big)$, which is 
asymptotically optimal in terms of classification risk, \emph{i.e.}  
$L_\infty(g_\infty^\star)  
= \inf_{g\text{ measurable } } L_\infty(g)$. 
Notice that for $x \in \rset_+^d\setminus\{0\}$, the angle  $\Theta(x)$ belongs to the positive orthant of the unit sphere, denoted by $S$ in the sequel.   
As a consequence, the optimal classifiers on extreme regions are based on indicator functions of truncated cones on the kind $\{\|x\| > t, \Theta(x) \in B \}$, where $B\subset S$, see  \autoref{fig:my_label}. We emphasize that  the labels provided by  such a classifier remain unchanged when rescaling the samples by a factor $\lambda \geq 1$ (\textit{i.e.} $g(x) = g(\Theta(x)) = g(\Theta(\lambda x)),  \forall x \in \{x, \|x \| \geq t \}$).
The angular structure of the optimal classifier $g_\infty^\star$  is the basis for the following  ERM strategy using the most extreme points of a  dataset.
 Let $\mathcal{G}_S$ be a class of angular classifiers defined on the sphere $S$ with finite VC dimension $V_{\mathcal{G}_S} < \infty $.
By extension, for any $x \in \rset^d_+$ and $g \in \mathcal{G}_S$,   $g(x) = g\big(\Theta(x)\big) \in \{-1, 1\}$. Given  $n$ training data  $\{(X_i, Y_i)\}_{i = 1}^n$ made of \textit{i.i.d} copies of $(X, Y)$, sorting the training observations by decreasing order of magnitude,  let $X_{(i)}$ (with corresponding sorted label $Y_{(i)}$) denote the $i$-th order statistic, \textit{i.e.} $\|X_{(1)}\| \geq \ldots \geq \|X_{(n)}\|$. The empirical risk for the $k$ largest observations $\widehat{L}_{k}(g)=\frac{1}{k}
  \sum_{i = 1}^{k} \mathbf{1}\{   Y_{(i)}\neq g(\Theta(X_{(i)}))\}  
$
is an empirical version of the risk $L_{t(k)}(g)$ as defined in~(\ref{eq:extremeRisk}) where $t(k)$ is a $(1 - k/n)$-quantile of the norm,   $\PP{\|X\| > t(k)} = k/n$. Selection of $k$ is a bias-variance compromise, see Appendix~\ref{sec:classif-supplem} for further discussion. 
The strategy promoted by \cite{jalalzai2018binary} is to use 
$ 
\widehat{g}_{k} = \arg\!\min_{g\in \mathcal{G}_S}\widehat{L}_{k}(g), 
$ 
for classification in the extreme region $\{ x \in \rset^{d}_+: \|x\| > t(k)\}$. 
The following result provides  guarantees  concerning  the excess risk of $\widehat g_k$ compared with the Bayes risk above level $t = t(k)$, $L_t^\star = \inf_{g \text{ measurable}}L_t(g) $.  
\begin{theorem}{(\cite{jalalzai2018binary}, Theorem~2)} \label{thm:main} 
If each class satisfies the regular variation assumption~(\ref{RV}), under an additional regularity assumption concerning the regression function $\eta(x) = \PP{Y = +1 \given x}$ 
(see Equation~(\ref{eq:uniformCV_regression}) in Appendix~\ref{sec:unifCV_app}), for $ \delta\in (0,1)$, $\forall n\geq 1$, it holds with probability larger than $1-\delta$ that
  \begin{multline*}
      L_{{t(k)}}(\widehat{g}_{k} )-L^\star_{{t(k)}}  \leq
 \frac{1}{\sqrt{k}} \Big(\sqrt{ 2(1- k/n) \log(2/\delta)}\ + 
    C\sqrt{V_{\mathcal{G}_S}\log(1/\delta)} \Big)   + \\   \frac{1}{k}\left(5 + 2\log(1/\delta) + \sqrt{ \log(1/\delta)}
    (C\sqrt{V_{\mathcal{G}_S}} +\sqrt 2)\right)
        + \left\{ \inf_{g\in \mathcal{G}_{S}}L_{{t(k)}}(g)-L^\star_{{t(k)}} \right\},
\end{multline*}
where $C$ is a universal constant. 
\end{theorem}
 In the present work we do \emph{not} assume that the baseline representation $X$ for text data satisfies the assumptions of Theorem~\ref{thm:main}. Instead, our goal  is 
is to render the latter theoretical framework applicable  by learning a representation which satisfies the regular variation condition given in~(\ref{RV}), hereafter referred as Condition~(\ref{RV}) which is the main assumption for  Theorem~\ref{thm:main} to hold. Our experiments demonstrate empirically  that enforcing  Condition~(\ref{RV})  is enough for our purposes,  namely improved classification and label preserving data augmentation,   see Appendix~\ref{sec:unifCV_app} for further discussion.

\section{Heavy-tailed Text Embeddings}
\label{sec:Contribution}
\subsection{Learning a heavy-tailed representation}
\label{subsec:classif}
We now introduce a novel algorithm  \emph{Learning a heavy-tailed representation} (${\fontfamily{qcr}\small\textbf{LHTR}}$) for  text data from high dimensional vectors as issued by pre-trained embeddings such as BERT. 
The idea behind 
is to modify the output $X$  of BERT
so that classification  in the  tail regions enjoys the statistical guarantees presented in Section~\ref{sec:MEVT}, while classification in the bulk (where many training points are available) can still be performed using standard models.
Stated otherwise, {\fontfamily{qcr}\small\textbf{LHTR}} increases the information carried by the resulting vector $Z = \varphi(X) \in \rset^{d^\prime}$ regarding the label $Y$  in  the tail regions of $Z$ 
in order to  improve  the performance of a downstream classifier. In addition  {\fontfamily{qcr}\small\textbf{LHTR}} is a building block of the data augmentation algorithm~\geneliex\ detailed in Section~\ref{sec:HTforDataAugment}. 
 {\fontfamily{qcr}\small\textbf{LHTR}} proceeds 
by training an encoding function $\varphi$  
in such  a way that \emph{(i)} the marginal distribution  $q(z)$ of the code $Z$ be close to a user-specified heavy tailed target distribution $p$ satisfying the regularity condition~(\ref{RV}); and \emph{(ii)} the classification loss of a multilayer perceptron trained on the code $Z$ be  small.

A major difference distinguishing {\fontfamily{qcr}\small\textbf{LHTR}} from existing auto-encoding schemes is that the target distribution on the latent space
is not chosen as a Gaussian distribution but as a heavy-tailed, regularly varying  one. A  workable example of such a target is provided in our experiments
(Section~\ref{sec:Exp}).
As the Bayes classifier  (\emph{i.e.}  the optimal one among all possible classifiers)  in the extreme region  has a  potentially  different structure from the  Bayes classifier on the bulk (recall from Section \ref{sec:MEVT} that the optimal classifier at infinity  depends on the angle $\Theta(x)$ only),
{\fontfamily{qcr}\small\textbf{LHTR}} trains two different classifiers, $g^{\text{ext}}$ on the extreme region of the latent space on the one hand, and $g^{\text{bulk}}$ on its complementary set on the other hand.
Given a high threshold~$t$, the extreme region of the latent space is defined as the set $\{z: \|z\|>t\}$. In practice, the threshold $t$ is chosen as an empirical quantile of order ($1-\kappa$) (for some small, fixed $\kappa$)  of the norm of encoded data $\|Z_i\| = \|\varphi(X_i)\|$.
The classifier trained  by {\fontfamily{qcr}\small\textbf{LHTR}} is thus of the kind
$ g(z) = g^{\text{ext}}(z)\un\{\|z\| >t \} + g^{\text{bulk}}(z)\un\{\|z\| \le t \}. $ 
  If the  downstream task is classification on the whole input space, in the end the bulk classifier $g^{\text{bulk}}$ may be replaced with  
  any other classifier $g^\prime$ trained on the original input data $X$ restricted to  the non-extreme samples (\textit{i.e.} $\{X_i,  \|\varphi(X_i)\| \leq t \}$). Indeed training
  $g^{\text{bulk}}$ only serves as an intermediate step to learn an  adequate
  representation~$\varphi$. 
\begin{remark}\label{rem:angularClassif}
  Recall from Section~\ref{sec:classifExtremes} that the optimal classifier in the extreme region as $t\to\infty$ depends on the angular component $\theta(x)$ only, or in other words, is scale invariant.
  One can thus reasonably expect the trained classifier $g^{\text{ext}}(z)$ to enjoy the same property. This scale invariance  is indeed verified in our experiments (see Sections~\ref{sec:Exp} and~\ref{sec:Exp_g}) and  is the starting point for our data augmentation algorithm in Section~\ref{sec:HTforDataAugment}. An alternative strategy would be to train an angular classifier, \emph{i.e.} to impose scale invariance. However in preliminary experiments (not shown here), the resulting classifier was less efficient and we decided against this option in view of the scale invariance and better performance of the unconstrained classifier. 
\end{remark}
The goal of \HTalgo\ is to minimize the weighted risk
\begin{equation*} 
\begin{aligned}
 R(\varphi, g^{\text{ext}}, g^{\text{bulk}}) =   &\rho_1  \PP{ Y \neq g^{\text{ext}}(Z), \|Z\|\ge t } + \\ &\rho_2\PP{ Y \neq g^{\text{bulk}}(Z), \|Z\|< t }
  + \\ &\rho_3 \mathfrak{D}(q(z), p(z))
\end{aligned}
\end{equation*}
where $Z = \varphi(X)$, $\mathfrak{D}$ is the Jensen-Shannon distance between the heavy tailed target distribution $p$ and the code distribution $q$, 
and  $\rho_1,\rho_2, \rho_3$ are  positive weights. 
Following common practice in the adversarial literature,  the Jensen-Shannon distance is  approached  (up to a constant term)
by the empirical proxy  $\widehat L(q , p)  = \sup_{D \in \Gamma} \widehat{ L} (q, p, D)$, with
 $ \widehat L(q , p, D)  = 
\frac{1}{m}\sum_{i=1}^m \log D (Z_i) + \log\big(1 - D (\tilde{Z_i})\big) $,  
where $\Gamma$ is a wide class of discriminant functions valued in $[0, 1]$, and where independent samples $Z_i, \tilde Z_i$ are respectively sampled from the target distribution and the code distribution $q$. Further details on adversarial learning are provided in Appendix~\ref{sec:adversarial_learning}.
The classifiers $g^{\text{ext}}$, $g^{\text{bulk}}$  are of the form $g^{\text{ext}} (z) = 2\un\{C^{\text{ext}}(z) > 1/2) - 1$, $g^{\text{bulk}}(z) = 2 \un\{C^{\text{bulk}}(z) > 1/2) - 1$ where
$C^{\text{ext}}, C^{\text{bulk}}$ are also discriminant functions valued in $[0,1]$. Following common practice, we shall refer to  $C^{\text{ext}}, C^{\text{bulk}}$ as classifiers as well. In the end, {\fontfamily{qcr}\small\textbf{LHTR}} solves the following min-max
problem
$\inf_{C^{\text{ext}}, C^{\text{bulk}}, \varphi}  \sup_{D}   \widehat{R}(\varphi, C^{\text{ext}}, C^{\text{bulk}}, D ) $
with
\begin{align*}
     \widehat{R}(\varphi, C^{\text{ext}}, C^{\text{bulk}}, D) =  \frac{\rho_1}{k} \sum_{i = 1}^k  \ell(Y_{(i)} ,  C^{\text{ext}}(Z_{(i)}) ) \  +
     \frac{\rho_2}{n - k} \sum_{i = k+1}^{n - k} \ell (Y_{(i)} ,  C^{\text{bulk}}(Z_{(i)}) ) + 
     \rho_3\, \hat L (q ,p, D ),
\end{align*}
 where $\{ Z_{(i)} = \varphi(X_{(i)}), i = 1,\ldots, n \}$ are the encoded observations with associated labels  $Y_{(i)}$ sorted by decreasing magnitude of $\|Z\|$ (\textit{i.e.} $\|Z_{(1)}\| \geq \cdots \geq \|Z_{(n)}\|$),
 $k = \lfloor  \kappa n \rfloor$ is the number of extreme samples among the $n$ encoded observations and
 $\ell(y, C(x)) = -( y \log C(x) + (1-y) \log(1-C(x)), y \in \{0,1\} $ is the negative log-likelihood of the discriminant function  $C(x) \in (0,1)$.  A summary of ${\fontfamily{qcr}\small\textbf{LHTR}}$ and an illustration of its workflow are provided in~Appendices~\ref{sec:workflows} and~\ref{sec:code}.

\subsection{A heavy-tailed representation for dataset augmentation}\label{sec:HTforDataAugment}
We now introduce {\fontfamily{qcr}\small\textbf{GENELIEX}} (Generating
Label Invariant sequences from Extremes), a data augmentation
algorithm, which relies on  the label invariance property under rescaling of the
classifier for the extremes 
learnt by
{\fontfamily{qcr}\small\textbf{LHTR}}. {\fontfamily{qcr}\small\textbf{GENELIEX}}
considers input sentences as sequences and follows the seq2seq approach \cite{google_seq,colombo2020guiding,chapuis2020hierarchical}. It trains a Transformer Decoder \cite{attention_is} $G^\text{ext}$ on the extreme regions.
For an input sequence $U = (u_{1},\dots, u_{T})$ of length $T$, represented as $X_U$ by BERT with latent code $Z = \varphi(X_U)$ lying in the extreme regions, {\fontfamily{qcr}\small\textbf{GENELIEX}} produces, through its decoder $G^\text{ext}$ $M$ sequences $U^\prime_{j}$ where $j \in \{1, \ldots, M\}$. The $M$ decoded sequences correspond to the codes $\{\lambda_j Z, j \in \{1, \ldots, M\}\}$ where $\lambda_j > 1$. To generate sequences, the decoder iteratively takes as input the previously generated word (the first word being a start symbol), updates its internal state, and returns the next word with the highest probability. This process is repeated until either the decoder generates a stop symbol or the length of the generated sequence reaches the maximum length~($T_{\max}$).
To train the decoder $G^\text{ext} : \rset^{d^\prime} \to \big[1,\ldots, |\mathcal{V}|\big]^{T_{\max}}$ where $\mathcal{V}$ is the vocabulary on the extreme regions,  {\fontfamily{qcr}\small\textbf{GENELIEX}} requires an additional dataset $\mathcal{D}_{g_n} = (U_1, \ldots,U_n)$  (not necessarily labeled) with associated representation \emph{via} BERT $(X_{U,1}, \ldots, X_{U,n})$. 
Learning is carried out by optimising the classical negative log-likelihood of individual tokens $\ell_{gen}$. The latter  is defined as $\ell_{gen}\big(U, G^\text{ext}(\varphi(X))\big) \defeq \sum_{t = 1}^{T_{\max}} \sum_{v \in \mathcal{V}} \un\{u_t = v\} \log\big(p_{v,t}\big)$, where $p_{v,t}$ is the probability predicted by $G^{\text{ext}}$ that the $t^{th}$ word is equal to $v$.  A detailed description of the training step of {\fontfamily{qcr}\small\textbf{GENELIEX}} is provided in Algorithm~\ref{alg:hydra} in Appendix \ref{sec:code}, see also Appendix~\ref{sec:workflows} for an illustrative diagram. 
\begin{remark}
  Note that the proposed method  only augments data on the extreme regions. A general data augmentation algorithm can be obtained by combining this approach with any other algorithm on the original input data $X$ whose latent code $Z = \varphi(X_U)$ does not lie in the extreme regions.
\end{remark}

\section{Experiments : Classification}
\label{sec:Exp}
In our experiments we work with the infinity norm. The proportion of extreme samples in the training step of \HTalgo\ is chosen as $\kappa = 1/4$. The threshold $t$ defining the extreme region
$\{\|x\|>t \}$ in the test set is  $ t = \| \tilde Z_{(\lfloor \kappa n \rfloor)}\|$  as returned by \HTalgo. 
We denote by $\mathcal{T}_{\text{test}}$
and $\mathcal{T}_{\text{train}}$ respectively the extreme test and train sets thus defined. 
Classifiers $C^\text{bulk}, C^\text{ext}$ involved in \HTalgo\ are  Multi Layer Perceptrons (MLP), see Appendix~\ref{additional_exp_classif} for a full description of the  architectures. \\
\textbf{Heavy-tailed distribution.}
The regularly varying target distribution is chosen  as a
multivariate logistic distribution with parameter $\delta =0.9$, refer to Appendix~\ref{logistic_Appendix} for details 
and  an illustration 
with various values of $\delta$.
This distribution  
is widely used in the context of extreme values analysis \cite{chiapino2019identifying, thomas2017anomaly,goix2016sparse} and differ from the classical logistic distribution.  
\subsection{Toy example: about \textbf{LHTR}}
\label{sec:toyExemple}

We start with a simple bivariate illustration of the heavy tailed
representation learnt by \HTalgo. Our
goal is to provide insight on how the learnt mapping $\varphi $ acts
on the input space and how the transformation affects the definition of
extremes (recall that extreme samples are defined as those samples
which norm exceeds an empirical quantile).
 \begin{figure}[ht]{}
 \centering
 \begin{subfigure}[t]{0.30\textwidth}
 \includegraphics[width=1\textwidth]{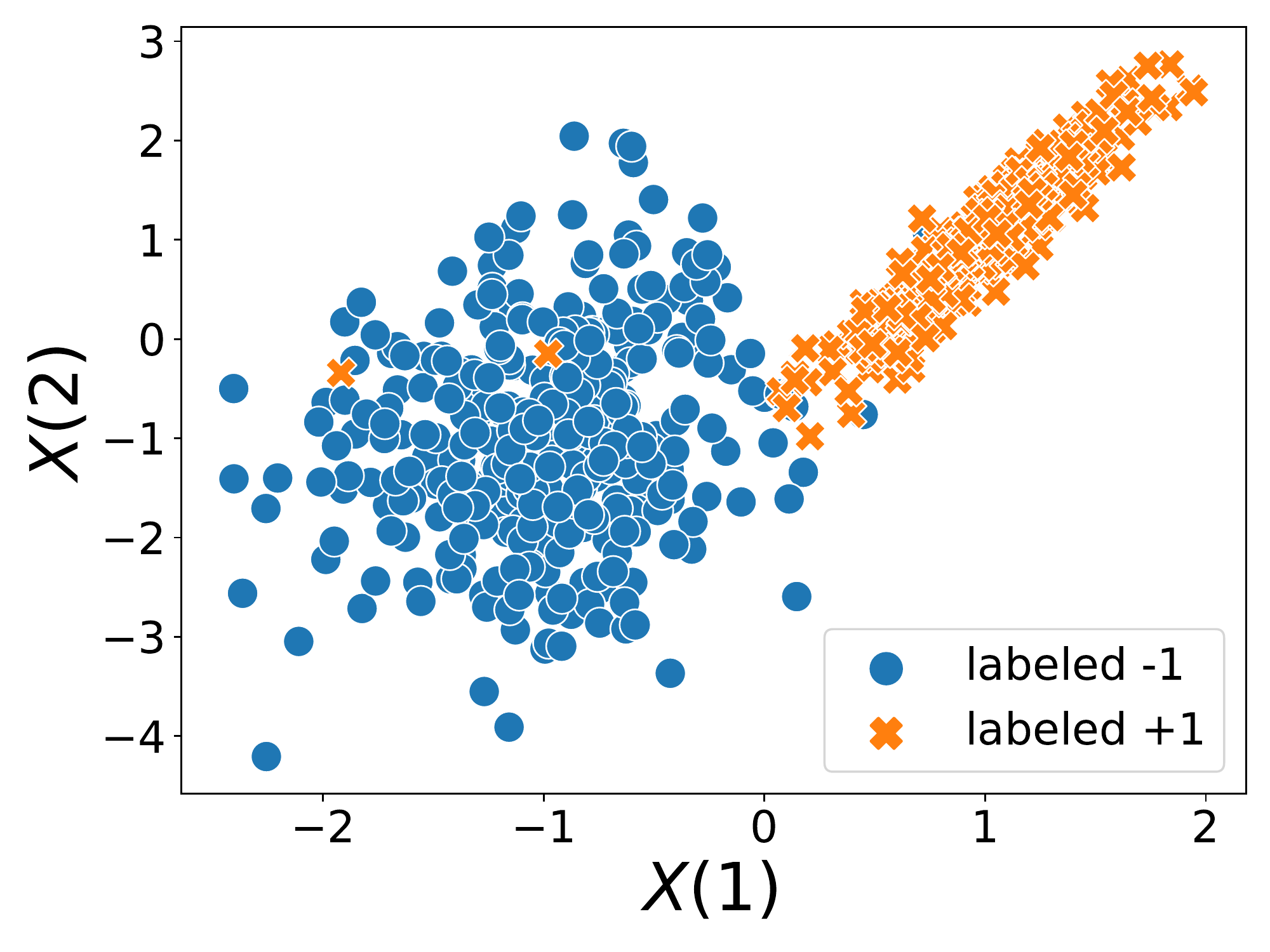}
 \caption{}
   \label{fig:classif}
 \end{subfigure}
 \begin{subfigure}[t]{0.30\textwidth}
   \includegraphics[width=\textwidth]{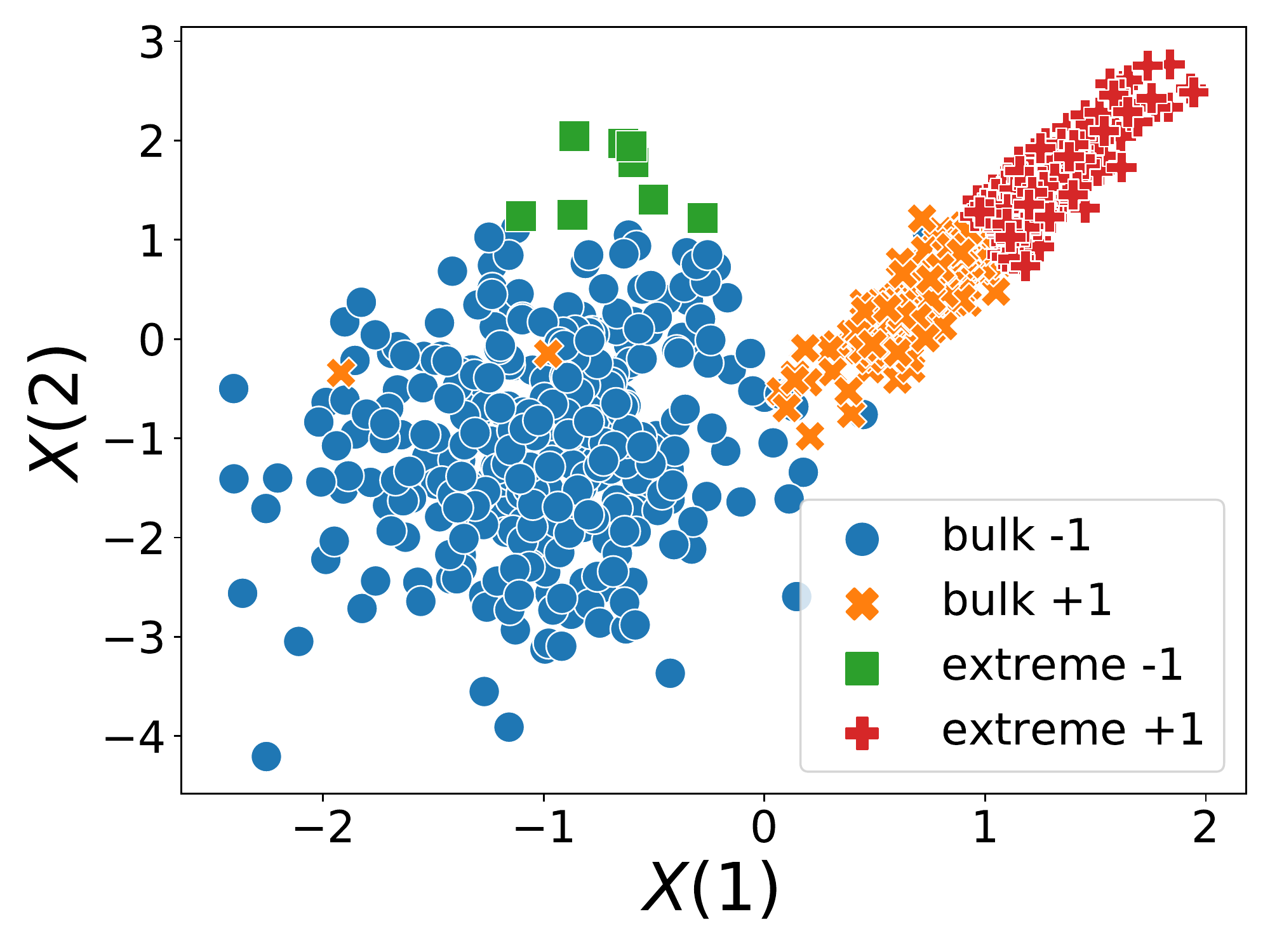}
   \caption{}
   \label{fig:classif_RawInput}
 \end{subfigure}\\
 \begin{subfigure}[t]{0.30\textwidth}
   \includegraphics[width=1\textwidth]{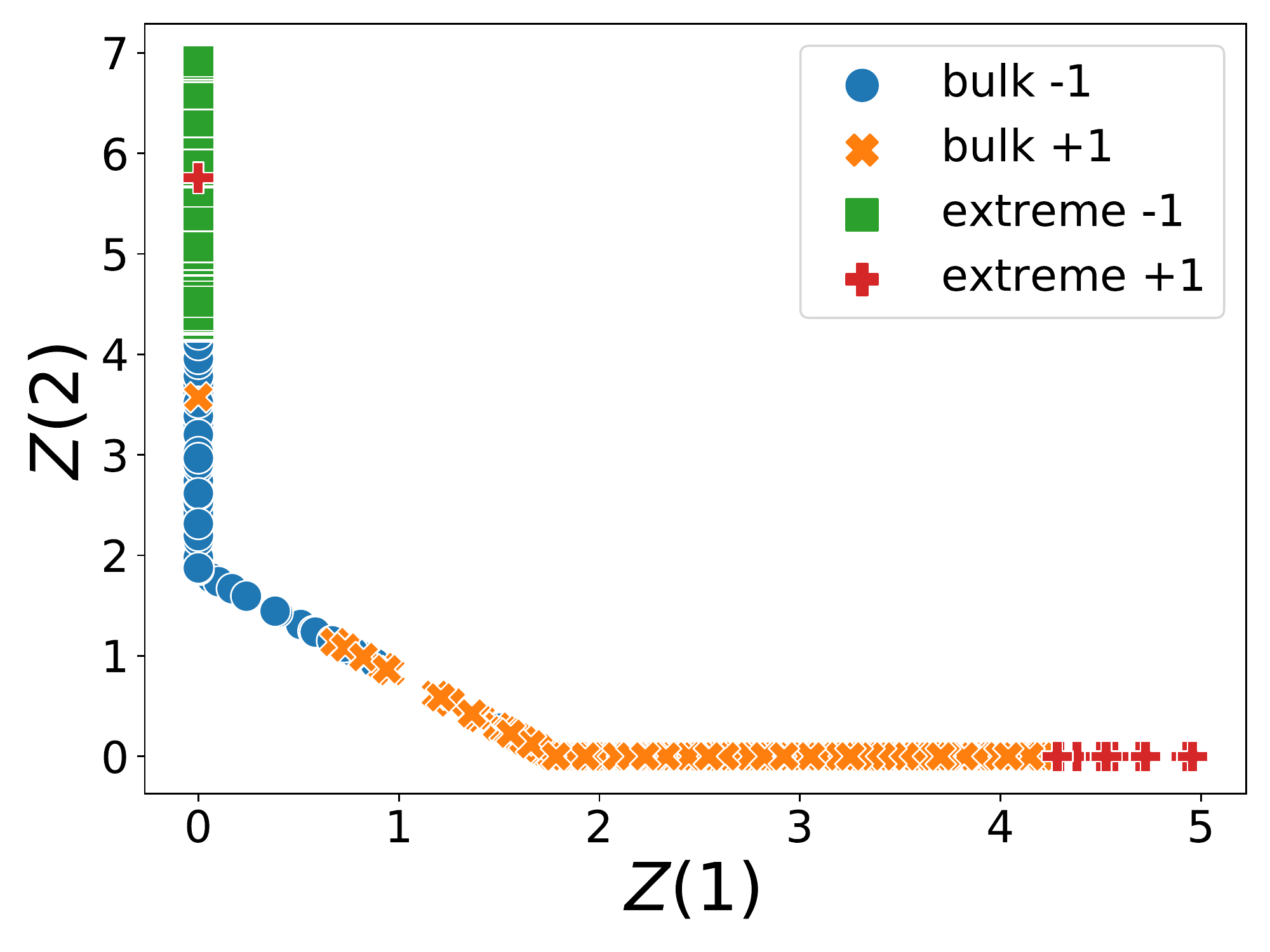}
   \caption{}
   \label{fig:classif_latent}
 \end{subfigure}
 \begin{subfigure}[t]{0.30\textwidth}
   \includegraphics[width=1\textwidth]{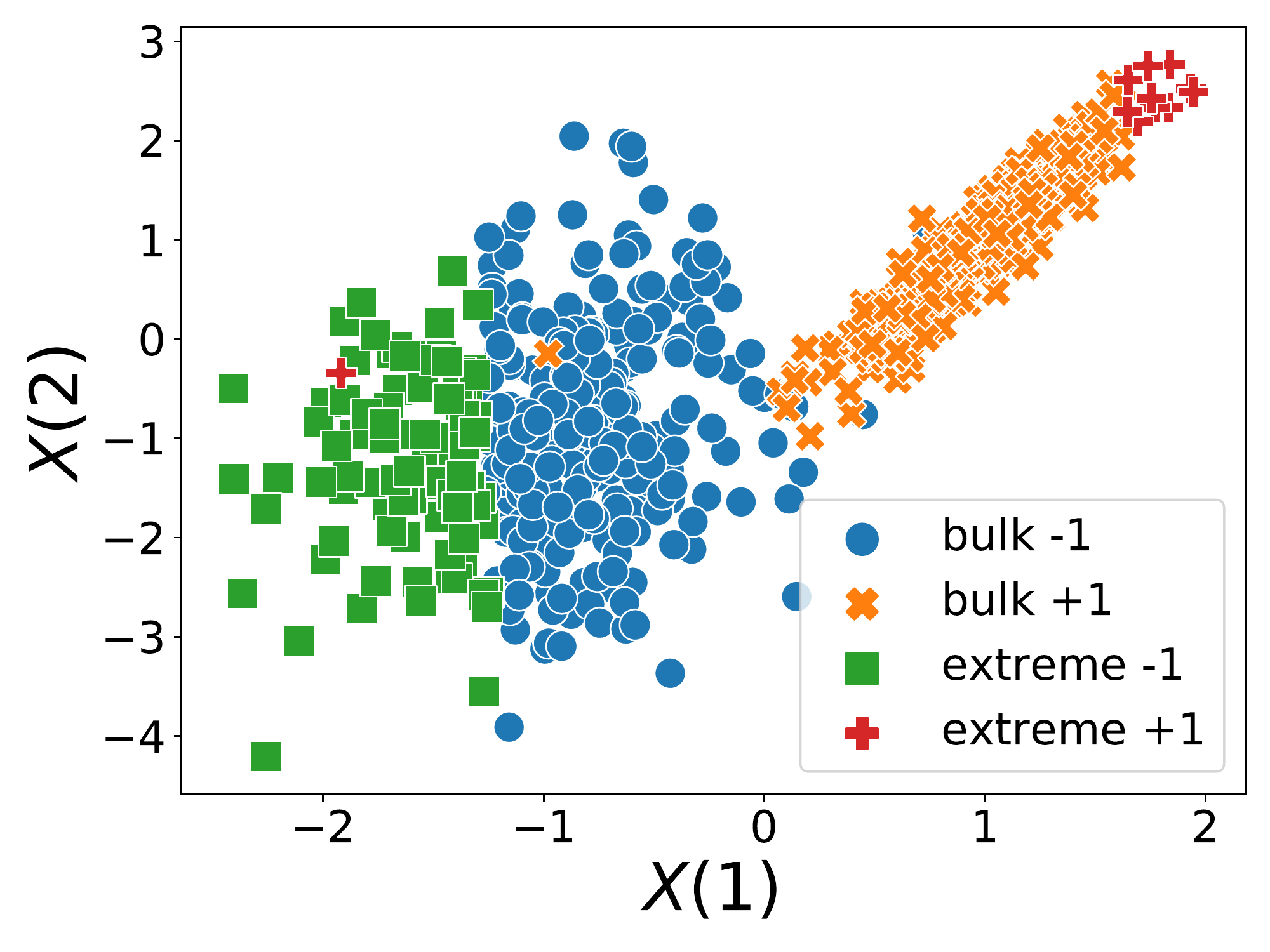}
   \caption{}
   \label{fig:Extreme_Orthrus}
 \end{subfigure}
 \caption{\autoref{fig:classif}: Bivariate samples $X_i$ 
   in the input space.
   \autoref{fig:classif_RawInput}: $X_i$'s in the input space  with extremes from each class  selected in the input space.
   \autoref{fig:classif_latent}:  Latent space representation  $Z_i = \varphi(X_i)$.  
   Extremes of each class are selected in the latent space.
  \autoref{fig:Extreme_Orthrus}:  $X_i$'s in the input space  with extremes from each class  selected in the latent  space.}
 \label{Fig:syntheticData_inputSpace}
 \end{figure} 
Labeled samples are simulated from  a Gaussian mixture distribution with two components of identical weight. The label indicates the component from which the point is generated.
\HTalgo\ is trained  on $2250$ examples and a  testing set of size $750$ is shown in Figure~\ref{Fig:syntheticData_inputSpace}. The testing samples
in the input space (Figure~\ref{fig:classif}) are mapped onto   the latent space \emph{via} $\varphi$ (Figure~\ref{fig:classif_latent}) 
In Figure~\ref{fig:classif_RawInput},  the extreme raw observations are selected according to their   norm after  a component-wise standardisation of $X_i$, 
 refer to Appendix~\ref{sec:classif-supplem} for details.  The  extreme threshold $t$ is chosen as the $75 \%$ empirical quantile of the norm on the training set in the input space.  Notice in the latter figure the class imbalance among extremes.
In Figure~\ref{fig:classif_latent}, extremes are  selected as the $25\%$  samples with the largest norm in the latent space. 
Figure~\ref{fig:Extreme_Orthrus}  is similar to Figure~\ref{fig:classif_RawInput} except for the selection of extremes which  is performed in the latent space as in Figure~\ref{fig:classif_latent}. On this toy example, the adversarial strategy appears to succeed in learning a code which distribution is close to the logistic target, as illustrated by the similarity between
 Figure~\ref{fig:classif_latent} and Figure~\ref{fig:logistic_alpha_09} in the supplementary. 
 In addition, the heavy tailed representation allows a more balanced  selection of extremes than the input representation.
 
\subsection{Application to positive \textit{vs.} negative classification of 
sequences}\label{sec:sentiment_analysisYelp}
\begin{figure}[t]{}
\centering
\includegraphics[width=0.8\textwidth]{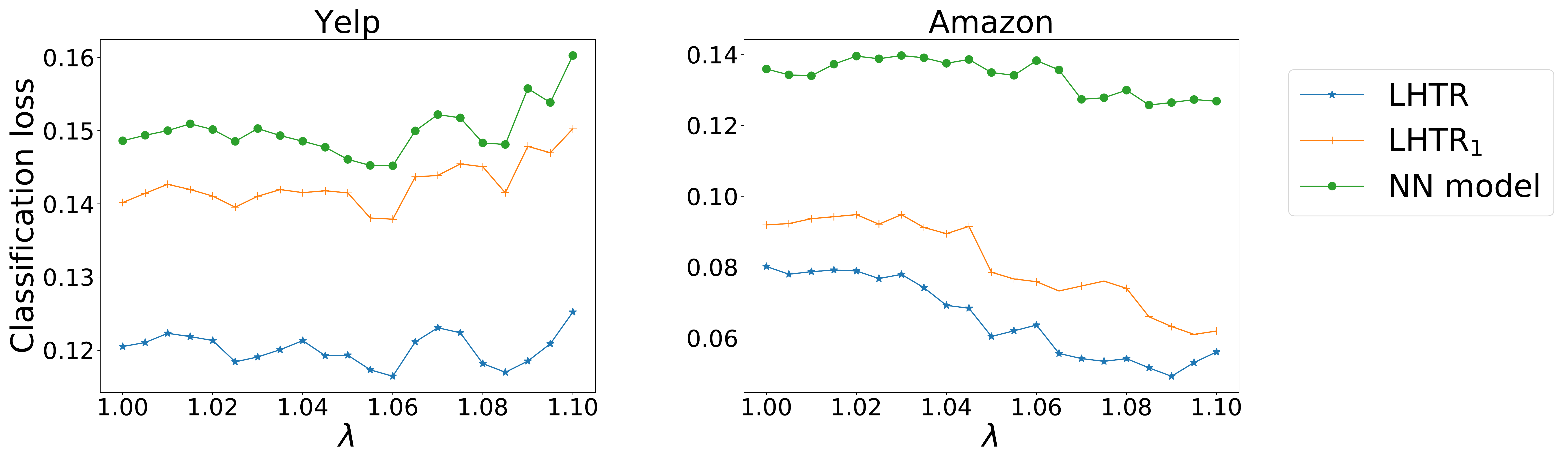}
\caption{Classification loss  of \HTalgo, \HTalgo$_1$ and  \NN\ 
  on the extreme test set $\{x \in \mathcal{T}, ||x|| \geq \lambda t \}$ for increasing values of $\lambda$ (X-axis), on \textit{Yelp} and 
  \textit{Amazon}.}
\label{fig:hamming_losses}
\end{figure}

In this section, we dissect \HTalgo\ to better understand the relative importance of: 
\emph{(i)}  working with a heavy-tailed representation, \emph{(ii)} training two independent classifiers: one dedicated to the bulk and the second one dedicated to the extremes. In addition, we verify experimentally that the latter  classifier is scale invariant, which is neither the case for  the former, nor for a classifier trained on BERT input. \\
\textbf{Experimental settings.} We compare the performance of three models. The baseline \NN\ is a MLP trained on BERT. The second model \HTalgo$_1$ is a variant of \HTalgo\ where a single MLP ($C$) is trained on the output of the encoder $\varphi$, using all the available data, both extreme and non extreme ones.  The third
model (\HTalgo) 
trains two separate MLP classifiers $C^{\text{ext}}$ and $C^{\text{bulk}}$  respectively  dedicated to the extreme and bulk regions of the learnt representation $\varphi$. All models take the same training inputs, use BERT embedding and their classifiers have identical structure,  see Appendix~\ref{sec:workflows} and~\ref{additional_exp_classif} for a summary of model workflows and  additional details concerning the network architectures. \\ 
Comparing \HTalgo$_1$ with \NN\ assesses the relevance of working with heavy-tailed embeddings.  Since \HTalgo$_1$  is obtained by using \HTalgo\   with $C^{\text{ext}} = C^{\text{bulk}}$, comparing  \HTalgo$_1$ with \HTalgo\ validates the use of two separate classifiers so that extremes are handled in a specific manner.  
As we make no claim concerning the usefulness of \HTalgo\ in the bulk, at the prediction step we suggest working with a combination of two models: \HTalgo\ with $C^{ext}$ for extreme samples and any other off-the-shelf ML tool for the remaining samples (\textit{e.g.} \NN).  \\
\textbf{Datasets.} In our experiments we rely on two large datasets from \textit{Amazon} (231k reviews)  \cite{mcauley2013hidden} and from \textit{Yelp} (1,450k reviews) \cite{yelp_1,yelp_2}. Reviews, (made of multiple sentences) with a rating greater than or equal to {\large \nicefrac{4\ }{\ 5}} are labeled as $+1$, while those  with a rating smaller or equal to {\large \nicefrac{2\ }{\ 5}} are labeled as $-1$. The gap in reviews' ratings is designed to avoid any overlap between labels of different contents.\\ 
\textbf{Results.} 
Figure~\ref{fig:hamming_losses} gathers the results obtained by the three considered classifiers on the tail regions of the  two datasets mentioned above. To illustrate the generalization ability of the proposed classifier in the extreme regions we consider nested  subsets of the extreme  test set $\mathcal{T}_{\text{test}}$,  $\mathcal{T^\lambda} = \{z \in \mathcal{T}_{\text{test}},  \|z\| \geq \lambda t\}$, $\lambda\ge 1$. For all factor $\lambda \geq 1 $, $\mathcal{T^\lambda} \subseteq \mathcal{T}_{\text{test}}$. The greater $\lambda$, the fewer the samples retained for evaluation and the greater their  norms. On both datasets, \HTalgo$_1$\ outperforms the baseline \NN.
This shows the improvement offered  by the heavy-tailed embedding on the extreme region. In addition,
\HTalgo$_1$ is in turn  largely   outperformed by the classifier \HTalgo, which  proves the importance of working with two separate classifiers.
The  performance of the proposed model respectively on the bulk region, tail region and overall, is reported in Table~\ref{tab:overall_results}, which shows that using a specific classifier dedicated to extremes improves the overall performance. 
\begin{table}[t]
\centering
   \begin{tabular}[h]{lcclccl}
   \hline 
   \multirow{2}{*}{Model}  &  \multicolumn{3}{c}{\textit{Amazon}} &   \multicolumn{3}{c}{\textit{Yelp}}  \\
 &Bulk  & Extreme & Overall & Bulk &Extreme  & Overall\\ \hline
   \NN\	&\textbf{0.085}		& 0.135 & 0.098 & \textbf{0.098}		& 0.148 & 0.111\\
\HTalgo$_1$		& 0.104		& 0.091 & 0.101 & 0.160  	& 0.139 &  0.155\\
\HTalgo			& 0.105		& \textbf{0.08}& 0.0988 & 0.162	& \textbf{0.1205} &  0.152\\\hline
\textbf{Proposed Model}	& \textbf{0.085} & \textbf{0.08}& \textbf{0.084} & \textbf{0.097}		& \textbf{0.1205}&\textbf{0.103} \\\hline
\end{tabular}
\caption{Classification losses on \textit{Amazon} and \textit{Yelp}. `Proposed Model' results from using \NN\ model for the bulk and \HTalgo\ for the extreme test sets. The extreme region contains $6.9$k samples for \textit{Amazon} and $6.1$k samples for  \textit{Yelp}, both corresponding roughly to $25\%$ of the whole test set size. 
}
\label{tab:overall_results}
\end{table}
\\\noindent\textbf{Scale invariance.} On all datasets,  the extreme classifier  $g^{\text{ext}} $ 
verifies Equation~(\ref{eq:invariance_dilation})     for each sample of the test set,
$    g^\text{ext}(\lambda Z) = g^\text{ext}(Z) $
 with $\lambda$ ranging from $1$ to $20$, demonstrating scale invariance of $g^\text{ext}$ on the extreme region. The  same experiments conducted both with \NN\ and a MLP classifier trained on BERT and \HTalgo$_1$ show label changes for  varying values of $\lambda$: none of them are scale invariant. Appendix~\ref{BERT_notRV} gathers additional experimental details. 
The scale invariance property will be exploited in the next section to perform label invariant generation.

\section{Experiments : Label Invariant Generation}
\label{sec:Exp_g}
\subsection{Experimental Setting}
\label{subsection:generation}
\textbf{Comparison with existing work.} We compare \geneliex\ with two state of the art methods for dataset augmentation, \citet{eda} and \citet{naacl_baseline}. Contrarily to these works 
which use heuristics and a synonym dictionary, \geneliex\ does not require any linguistic resource. 
To ensure that the improvement brought by  \geneliex\ is not only due to BERT, we have updated the method in \cite{naacl_baseline} with a BERT language model 
(see Appendix \ref{additional_exp_generation} for details and Table~\ref{tab:generation_experiments} for hyperparameters).\\
\textbf{Evaluation Metrics.} Automatic evaluation of generative models for text is still an open research problem. We rely both on perceptive evaluation and automatic measures to evaluate our model through four criteria (\textbf{C1}, \textbf{\textbf{C2}}, \textbf{C3},\textbf{\textbf{C4}}). \textbf{C1} measures Cohesion \cite{cohesion} (\emph{Are the generated sequences 
grammatically and semantically consistent?}). \textbf{\textbf{C2}} (named Sent. in Table \ref{tab:qualitative_evaluation}) evaluates label conservation (\emph{Does the expressed sentiment in the generated sequence 
match the sentiment of the input sequence?}). \textbf{C3} measures the diversity \cite{diversity} (corresponding to dist1 or dist2 in Table \ref{tab:qualitative_evaluation}\footnote{dist\,$n$ is obtained by calculating the number of distinct $n$-grams divided by the total number of generated tokens to avoid favoring long sequences
.}) of the sequences 
(\emph{Does the augmented dataset contain diverse sequences?}). Augmenting the training set with very diverse sequences 
can lead to better classification performance. \textbf{\textbf{C4}} measures the improvement in terms of F1 score when training a classifier (fastText \cite{fastText}) on the augmented training set (\emph{Does the augmented dataset improve classification performance?}). \\
\textbf{Datasets.} \geneliex\ is evaluated on two datasets, a medium and a large one (see \cite{regime}) which respectively contains 1k and 10k labeled samples. In both cases, we have access to $\mathcal{D}_{g_{n}}$ a dataset of 80k unlabeled samples. Datasets are randomly  sampled from \textit{Amazon} and \textit{Yelp}. \\
\textbf{Experiment description.} We augment extreme regions of each dataset according to three algorithms: \geneliex\ (with scaling factor $\lambda$ ranging from 1 to 1.5)
, \citet{naacl_baseline}, and  \citet{eda}. For each train set's sequence 
considered as extreme, $10$ new sequences 
are generated using each algorithm.  Appendix~\ref{additional_exp_generation} gathers further details. For experiment \textbf{C4} the test set contains $10^4$ sequences. 
\subsection{Results}
\textbf{Automatic measures.}
The results of \textbf{C3} and \textbf{C4} evaluation are reported  in Table~\ref{tab:quant_eval}. Augmented data with \geneliex\ are more diverse than the one augmented with \citet{naacl_baseline} and \citet{eda}. The F1-score with dataset augmentation performed by  \geneliex\ outperforms the aforementioned methods on Amazon in medium and large dataset and on Yelp for the medium dataset. It equals state of the art performances on Yelp for the large dataset. As expected, for all three algorithms, the benefits of data augmentation decrease as the original training dataset size increases.  
Interestingly, we observe a strong correlation between more diverse 
sequences in the extreme regions and higher F1 score: the more diverse the augmented dataset, the higher the F1 score. More diverse sequences are thus more likely to lead to better improvement on downstream tasks (\emph{e.g.} classification). 
\begin{table}[t!]
\centering
   \begin{tabular}[h]{lcccccccc}
   \hline 
   \multirow{3}{*}{Model} & \multicolumn{4}{c}{\textit{Amazon}} & \multicolumn{4}{c}{\textit{Yelp}}     \\
  &  \multicolumn{2}{c}{Medium} &   \multicolumn{2}{c}{Large}  &  \multicolumn{2}{c}{Medium} &   \multicolumn{2}{c}{Large}  \\
 &\small{F1} &\small dist1/dist2 &\small F1 &\small dist1/dist2  &\small{F1} &\small dist1/dist2 &\small F1 &\small dist1/dist2 \\ \hline
    \small Raw Data &\small 84.0  &\small X &\small 93.3 &\small X &\small 86.7 &\small X &\small 94.1 &\small X \\
   \small \citet{naacl_baseline} & \small85.0 &\small 0.10/0.47 &\small 92.9 &\small 0.14/0.53 & \small87.0&\small 0.15/0.53  &\small 94.0 &\small 0.14/0.58 \\
   \small \citet{eda} &\small 85.2&\small 0.11/0.50 &\small 93.2 &\small 0.14/0.54 &\small 87.0 &\small 0.15/0.52 &\small \textbf{94.2} &\small  \textbf{0.16}/0.59 \\
  \small \geneliex\ & \small \textbf{86.3} &\small \textbf{0.14/0.52} &\small \textbf{94.0} &\small \textbf{0.18/0.58} & \small \textbf{88.4} &\small \textbf{0.18/0.62} &\small \textbf{94.2} &\small \textbf{0.16/0.60} \\
   \hline 
\end{tabular}
\caption{Quantitative Evaluation. Algorithms are compared according to \textbf{C3} and \textbf{C4}. dist1 and dist2 respectively stand for distinct~1 and~2, it measures the diversity of new sequences 
in terms of unigrams and bigrams. F1  is the F1-score for FastText classifier trained on an augmented labelled training set.}
\label{tab:quant_eval}
\end{table}
\begin{table}[t]
\centering
   \begin{tabular}[h]{lcccc}
   \hline 
   \multirow{2}{*}{Model}
  &  \multicolumn{2}{c}{\textit{Amazon}} &   \multicolumn{2}{c}{\textit{Yelp}}  \\
 &\small{Sent.} &\small Cohesion &\small Sent. &\small Cohesion \\ \hline
    \small Raw Data &\small 83.6  &\small 78.3 &\small 80.6 &\small 0.71  \\
   \small \citet{naacl_baseline} & \small \textbf{80.0}  &\small \textbf{84.2} &\small 82.9 &\small 0.72 \\
   \small \citet{eda} &\small 69.0 &\small 67.4 &\small 80.0 &\small 0.60 \\
  \small \geneliex\ & \small 78.4 &\small 73.2 &\small \textbf{85.7} &\small \textbf{0.77} \\ 
   \hline 
\end{tabular}
\caption{Qualitative evaluation with three turkers. Sent. stands for sentiment label preservation. The Krippendorff Alpha for Amazon is $\alpha=0.28$ on the sentiment classification and $\alpha=0.20$ for cohesion. The Krippendorff Alpha for Yelp is $\alpha=0.57$ on the sentiment classification and $\alpha=0.48$ for cohesion. }
\label{tab:qualitative_evaluation}
\end{table}\\
\textbf{Perceptive Measures.}
To evaluate \textbf{C1}, \textbf{C2}, three turkers were asked to annotate the cohesion and the sentiment of $100$ generated sequences 
for each algorithm and for the raw data. F1 scores of this evaluation are reported in Table \ref{tab:qualitative_evaluation}.
Grammar evaluation confirms the findings of \cite{eda} showing that random swaps and deletions do not always maintain the cohesion of the sequence. 
In contrast,  \geneliex\ and \citet{naacl_baseline}, using vectorial representations, produce more coherent sequences. 
Concerning  sentiment label preservation, on Yelp,  \geneliex\ achieves the highest score which confirms the observed improvement reported in Table~\ref{tab:quant_eval}. On Amazon, turker annotations with data from \geneliex\ obtain a lower F1-score than from \citet{naacl_baseline}. This does not correlate with results in Table~\ref{tab:quant_eval} and may be explained by a lower Krippendorff Alpha\footnote{measure of inter-rater reliability in $[0,1]$: $0$ is perfect disagreement and $1$ is perfect agreement.} on Amazon ($\alpha=0.20$)  than on Yelp ($\alpha = 0.57$).

\section{Broader Impact}
In this work, we propose a method resulting in heavy-tailed text embeddings. As we make no assumption on the nature of the input data, the suggested method is not limited to textual data and can be extended to any type of modality (\textit{e.g.} audio, video, images).
A classifier, trained on aforementioned embedding is dilation invariant (see Equation~\ref{eq:invariance_dilation}) on the extreme region. A dilation invariant classifier enables better generalization for new samples falling out of the training envelop.  For critical application ranging from web content filtering (\textit{e.g.} spam~\cite{guzella2009review}, hate speech  detection~\cite{davidson2017automated}, fake news~\cite{perez2017automatic} or multi-modal classification \cite{garcia2019token}) to medical case reports to court decisions it is crucial to build classifiers with lower generalization error. The scale invariance property can also be exploited to automatically augment a small dataset on its extreme region. For application where data collection requires a huge effort both in time and cost (\textit{e.g.} industrial factory design, classification for rare language~\cite{scare}), beyond industrial aspect, active learning problems involving heavy-tailed data may highly benefit from our data augmentation approach. 

\section{Acknowledgement}
Anne Sabourin was partly supported by the Chaire \textit{Stress testing} from Ecole Polytechnique and BNP Paribas. Concerning Eric Gaussier, this project partly fits within the MIAI project (ANR-19-P3IA-0003).
\bibliography{main_icml.bbl}
\bibliographystyle{plainnat}
\appendix
\onecolumn
\begin{center}
  {\large {\bf\textsc{supplementary material : heavy-tailed representations, text polarity classification \& data augmentation}}}
\end{center}

\section{Models}\label{sec:Model_sup} 
\subsection{Background on Adversarial Learning}\label{sec:adversarial_learning}
Adversarial networks, introduced in \cite{goodfellow2014generative}, form a system where two neural networks are competing. 
A first model $G$, called the generator, generates  samples as close as possible to the input dataset. A second model $D$, called the discriminator, aims at distinguishing samples produced by the generator from the input dataset. 
The goal of the generator is to maximize the probability of  the discriminator making a mistake. Hence, if $P_\text{input}$ is the distribution of the input dataset then the adversarial network intends to minimize the distance (as measured by the Jensen-Shannon divergence) between the distribution of the generated data $P_G$ and $P_\text{input}$. In short, the problem is a minmax game with value function $V(D, G)$
\begin{align*}
    \min_{G}\max_{D} V(D, G) =  &\mathbb{E}_{x \sim P_{\text{input}} }[\log D(x)] + \mathbb{E}_{z \sim P_{G}}[\log\big(1 - D(G(z))\big)].
\end{align*}
Auto-encoders and derivatives \cite{Goodfellow-et-al-2016, laforgue2018autoencoding, fard2018deep} form a subclass of neural networks whose purpose is to build a suitable representation by learning encoding and decoding functions which capture the core properties of the input data.
An adversarial auto-encoder (see \cite{makhzani2015adversarial}) is a specific kind of auto-encoders where the encoder plays the role of the generator of an adversarial network. Thus the latent code is forced to follow a given distribution while containing information relevant to reconstructing the  input. In the remaining of this paper, a similar adversarial encoder constrains the encoded representation to be heavy-tailed. 

\subsection{Models Overview}\label{sec:workflows} 
\begin{figure}[h]{}
 \begin{subfigure}[t]{0.49\textwidth}
 \includegraphics[width=1\textwidth]{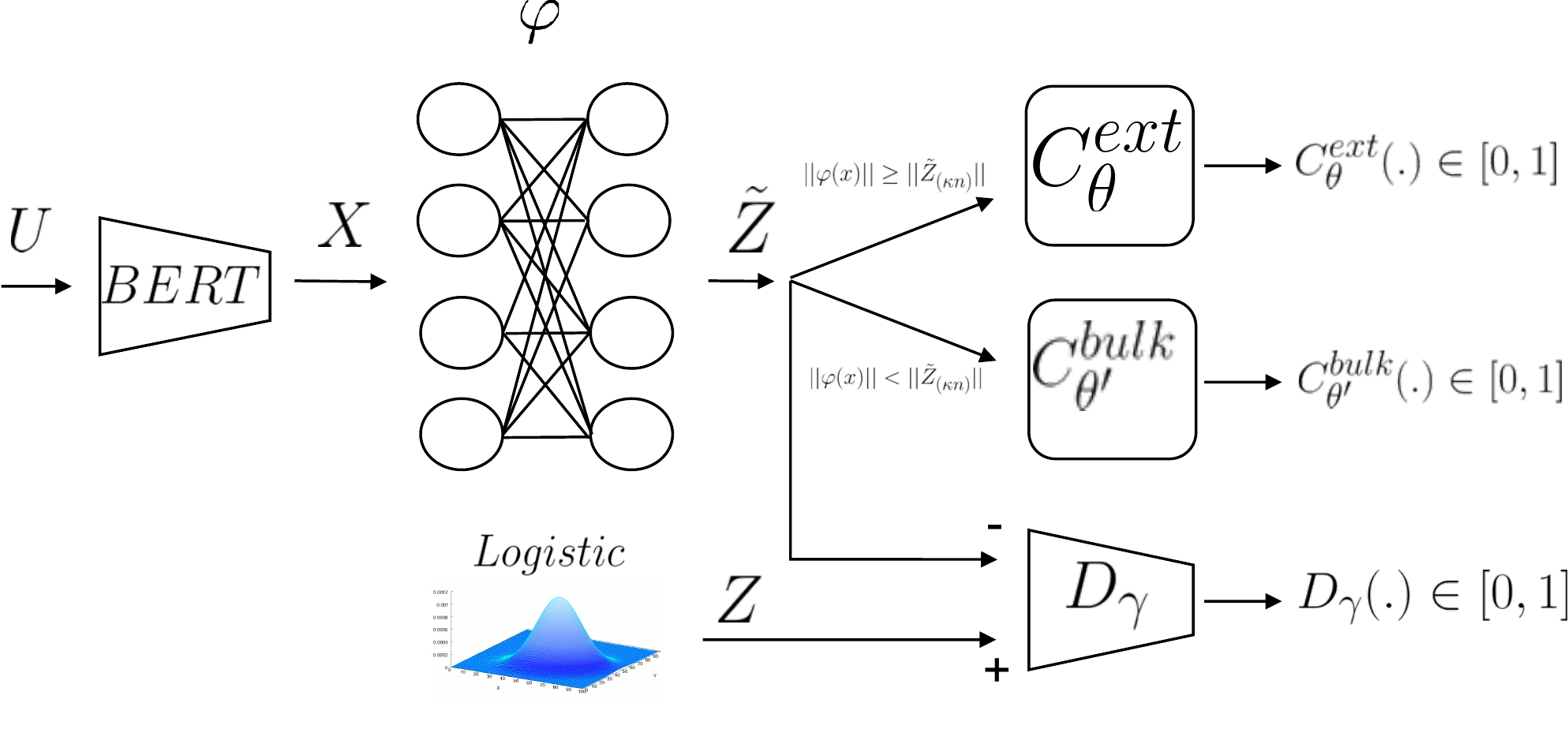}
 \caption{}
   \label{fig:two_heads}
 \end{subfigure}
 \begin{subfigure}[t]{0.49\textwidth}
   \includegraphics[width=\textwidth]{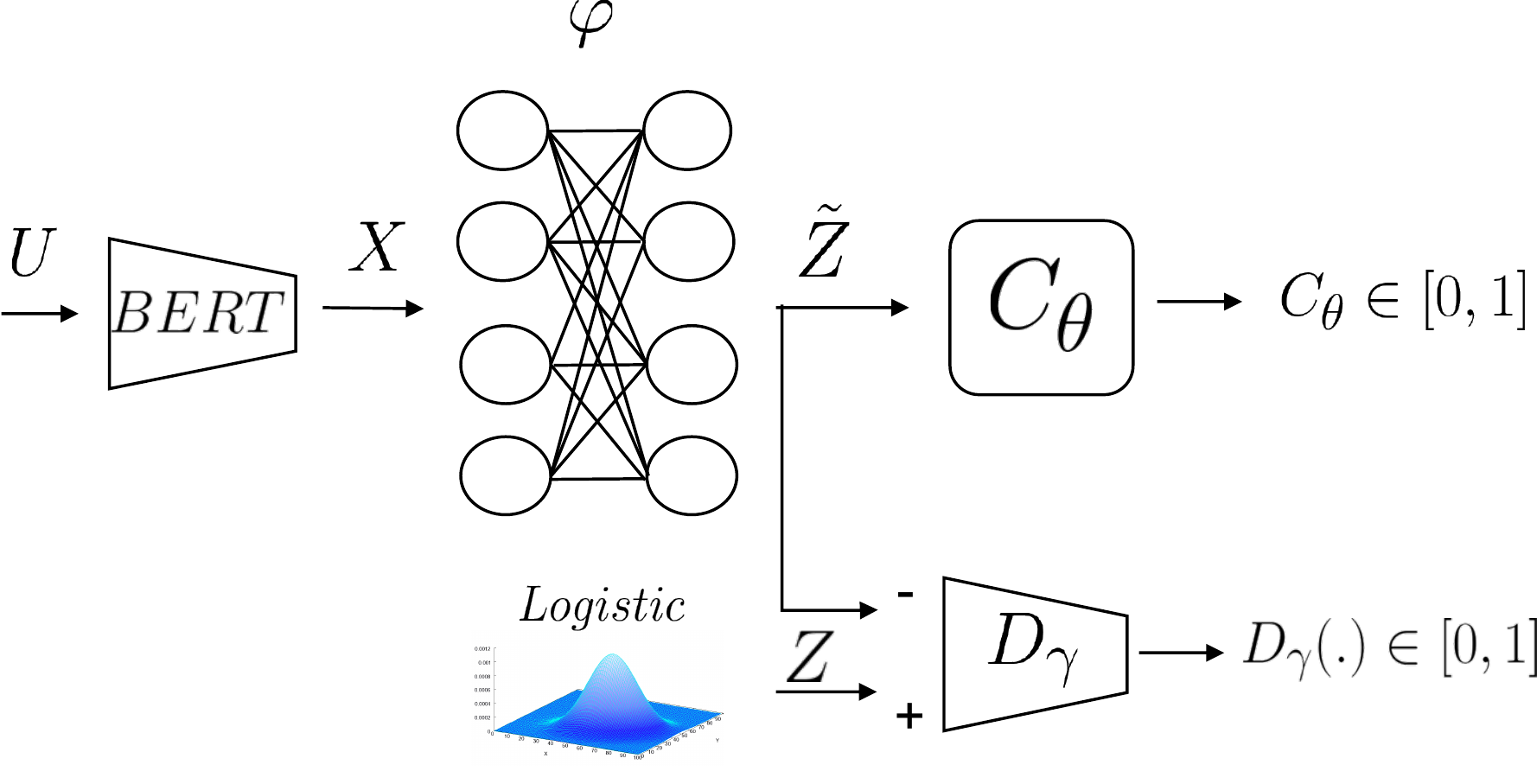}
   \caption{}
   \label{fig:one_head}
 \end{subfigure}
 \begin{subfigure}[t]{0.49\textwidth}
   \includegraphics[width=1\textwidth]{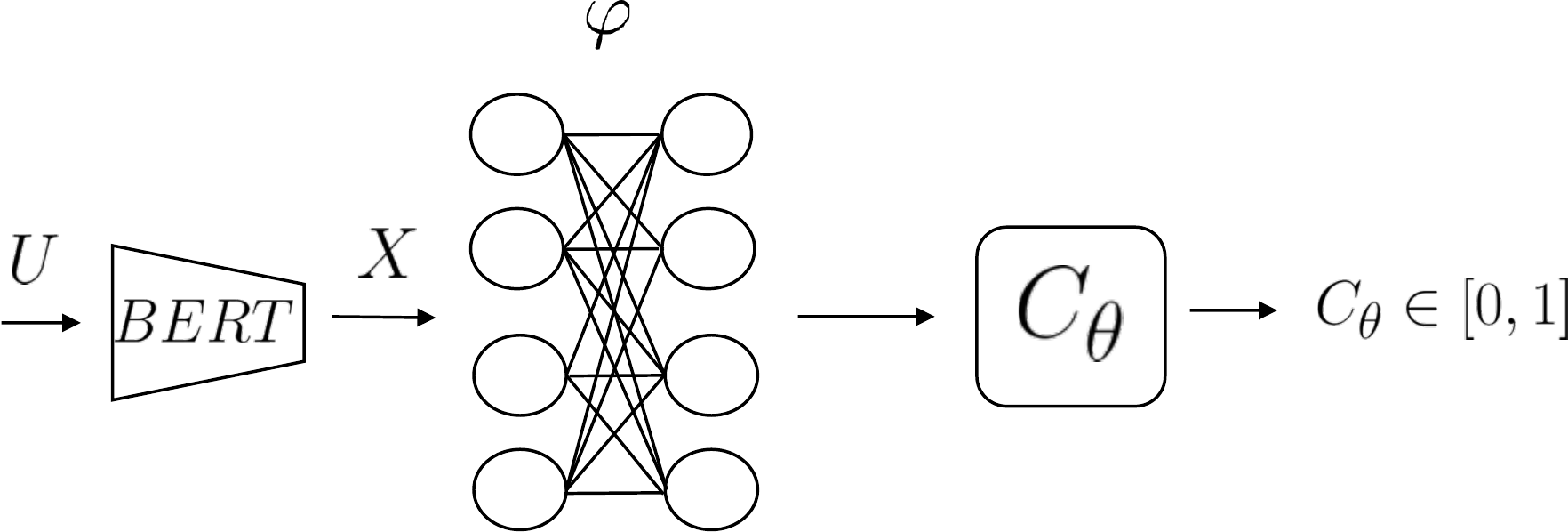}
   \caption{}
   \label{fig:baseline}
 \end{subfigure}
 \hfill
 \begin{subfigure}[t]{0.49\textwidth}
   \includegraphics[width=1\textwidth]{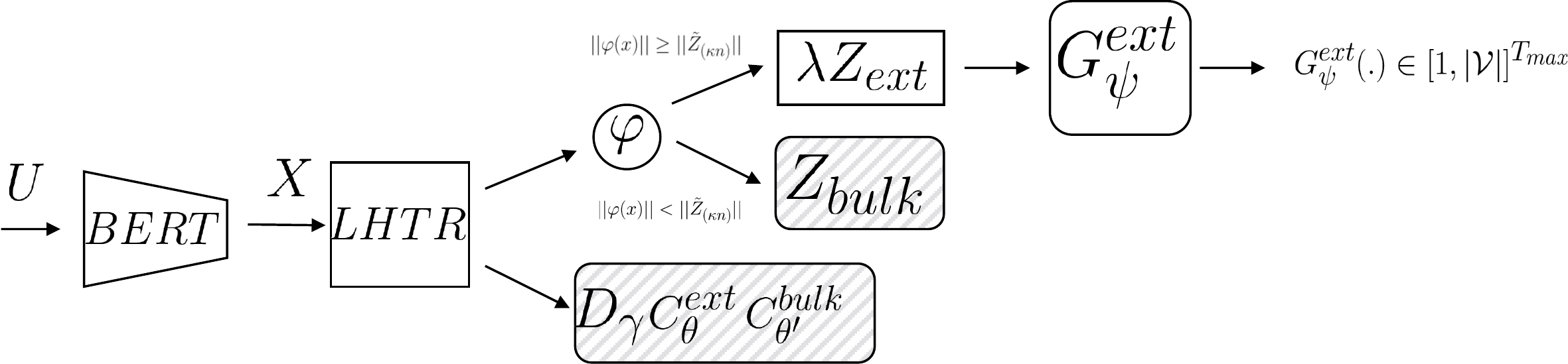}
   \caption{}
   \label{fig:generation}
 \end{subfigure}
 \caption{Illustrative pipelines.}
 \label{Fig:schema}
 \end{figure}

Figure~\ref{Fig:schema} provides an overview of the different algorithms proposed  in the paper. Figure~\ref{fig:two_heads} describes the pipeline for {\fontfamily{qcr}\small\textbf{LHTR}} detailed in Algorithm~\ref{alg:Orthrus}. Figure~\ref{fig:one_head} describes the pipeline for the comparative baseline~\HTalgo$_1$  where $C^\text{ext} = C^\text{bulk}$. Figure~\ref{fig:baseline} illustrates the pipeline for the baseline classifier trained on BERT. Figure~\ref{fig:generation} describes  {\fontfamily{qcr}\small\textbf{GENELIEX}} described in Algorithm~\ref{alg:hydra}, note that the hatched components are inherited from \HTalgo~ and are not used in the workflow.

\subsection{\HTalgo\ and \geneliex\ algorithm}\label{sec:code} 
This subsection provides detailed algorithm for both models \HTalgo\ and \geneliex.
\begin{algorithm}[t]
\caption{LHTR}
\begin{algorithmic}
\INPUT  Weighting coef. $\rho_1, \rho_2, \rho_3>0$, Training dataset $\mathcal{D}_n=\{(X_1,Y_1),\; \ldots,\; (X_n,Y_n)\}$, batch size $m$, proportion of extremes $\kappa$, heavy tailed prior $P_Z$. 
\Initialize parameters $(\tau, \theta,\theta', \gamma)$ of the encoder $\varphi_\tau$, classifiers $C_\theta^{ext}$, $C_{\theta^\prime}^{bulk}$ and  discriminator $D_\gamma$
\Optimization
\While{$(\tau, \theta, \theta^\prime,\gamma)$ not converged}
\Statex Sample $\{(X_1, Y_1) \ldots, (X_m, Y_m)\}$ from $\mathcal{D}_n$ 
and define $\tilde Z_i = \varphi(X_i), \, i\le m$. 
\Statex Sample $\{Z_1, \ldots, Z_m\}$ from the prior $P_Z$.
\Statex Update $\gamma$ by ascending:\\
\Statex 
 $$ \frac{\rho_3}{m}\sum_{i=1}^m \log D_\gamma(Z_i) + \log(1 - D_\gamma(\tilde{Z_i})).$$
\Statex Sort $\{\tilde{Z_i}\}_{i \in \{1,\ldots, m\}}$ by decreasing order of magnitude $||\tilde{Z}_{(1)}||\geq \ldots \geq ||\tilde{Z}_{(m)}||.$
\Statex Update 
$\theta$ by descending:\\
\Statex 
$$ \mathcal{L}^\text{ext} (\theta,\tau) \defeq \frac{\rho_1}{\lfloor\kappa m\rfloor}\sum_{i=1}^{\lfloor{\kappa m}\rfloor} \ell\big(Y_{(i)} ,  C^\text{ext}_\theta(\tilde{Z}_{(i)})\big).$$
\Statex Update 
$\theta^\prime$ by descending:\\
\Statex
$$\mathcal{L}^\text{bulk}(\theta^\prime,\tau) \defeq \frac{\rho_2}{m - \lfloor{\kappa m}\rfloor}\sum_{i=\lfloor{\kappa m}\rfloor+1}^m \ell\big( Y_{(i)} ,  C_{\theta^\prime}^\text{bulk}(\tilde{Z}_{(i)})\big).$$
\Statex
\Statex Update $\tau$ by descending: 
\Statex
$$ \frac{1}{m}\sum_{i=1}^m 
- \rho_3 \log D_\gamma(\tilde{Z}_{i})
+ \mathcal{L}^\text{ext}(\theta,\tau) + \mathcal{L}^\text{bulk}(\theta^\prime,\tau).
$$
\EndWhile
\Statex Compute $\{\tilde{Z_i}\}_{i \in \{1,\ldots, n\}} = {\varphi(X_i)}_{i \in \{1,\ldots, n\}}$
\Statex Sort $\{\tilde{Z_i}\}_{i \in \{1,\ldots, n\}}$ by decreasing order of magnitude $||\tilde{Z}_{(1)}||\geq \ldots ||\tilde{Z}_{(\lfloor{\kappa n}\rfloor)}||\geq \ldots \geq ||\tilde{Z}_{(n)}||.$
\OUTPUT encoder $\varphi$, classifiers $C^\text{ext}$ for $\{x : ||\varphi(x)|| \geq  t:=||\tilde{Z}_{(\lfloor{\kappa n}\rfloor)}||  \}$ and $C^\text{bulk}$ on the complementary set.
\end{algorithmic}
\label{alg:Orthrus}
\end{algorithm}
\begin{algorithm}[H]
\caption{GENELIEX: training step}
\begin{algorithmic}
\INPUT 
input of LHTR, 
$\mathcal{D}_{g_{n}}=\{U_{1},\; \ldots,\; U_{n}\}$
\Initialize parameters of $\varphi_\tau$, $C_\theta^{\text{ext}}$, $C_{\theta^\prime}^{\text{bulk}}$, $D_\gamma$ and decoder $G_\psi^{\text{ext}}$ 
\Optimization  
\Statex  $\varphi$, $C^{\text{ext}}$, $C^{\text{bulk}}$ = LHTR($\rho_1, \rho_2, \rho_3$, $\mathcal{D}_{n},\kappa, m$)
\While {$\psi$ not converged} 
\Statex Sample $\{U_{1} \ldots, U_{m}\}$ from the training set $\mathcal{D}_{g_n}$ and define
$\tilde{Z_i} = \varphi( X_{U,i})$ for $i \in \{1,\ldots, m\}$.
\Statex Sort $\{\tilde{Z_i}\}_{i \in \{1,\ldots, m\}}$ by decreasing order of magnitude
$\|\tilde{Z}_{(1)}\| \geq \ldots \geq \|\tilde{Z}_{(m)}\|.$
\Statex Update 
$\psi$ by descending:
$$\mathcal{L}_g^\text{ext}(\psi) \defeq \frac{\rho_1}{\lfloor{\kappa m}\rfloor}\sum_{i=1}^{\lfloor{\kappa m}\rfloor} \ell_{gen.} \big(U_{(i)}, G^\text{ext}_\psi(\tilde{Z}_{(i)})\big).$$
%
\EndWhile
\Statex Compute $\{\tilde{Z_i}\}_{i \in \{1,\ldots, n\}} = {\varphi(X_i)}_{i \in \{1,\ldots, n\}}$
\Statex Sort $\{\tilde{Z_i}\}_{i \in \{1,\ldots, n\}}$ by decreasing order of magnitude $\|\tilde{Z}_{(1)}\|\geq \ldots \|\tilde{Z}_{(k)}\|\geq \ldots \geq \|\tilde{Z}_{(n)}\|.$
\OUTPUT encoder $\varphi$, decoder $G^\text{ext}$ applicable on the region $\{x : \|\varphi(x)\| \geq \|\tilde{Z}_{(\lfloor{\kappa n}\rfloor)}\|  \}$ 
\end{algorithmic}
\label{alg:hydra}
\end{algorithm}

\section{Extreme Value Analysis: additional material}
\label{sec:classif-supplem}

\subsection{Choice of~k}  
To the best of our knowledge, selection of $k$  in extreme value analysis (in particular in  Algorithm~\ref{alg:Orthrus} and Algorithm~\ref{alg:hydra}) is still a vivid problem in EVT for which no absolute answer exists. As $k$ gets large the number of extreme points increases including samples which are not large enough and deviates from the asymptotic distribution of extremes. Smaller values of $k$ increase the variance of the classifier/generator. This bias-variance trade-off is beyond the scope of this paper.

\subsection{Preliminary standardization for selecting extreme samples} 
In Figure~\ref{fig:classif_RawInput} selecting the extreme samples on the input space is not a straightforward step as the two components of the vector are not on the same scale, componentwise standardisation is a natural and necessary preliminary step. Following common practice in multivariate extreme value analysis it was decided to standardise the input data $(X_i)_{i \in \{1,\ldots, n\} }$ by applying the rank-transformation: 
$$\widehat{T}(x) =\bigg(1 / \Big(1 - \widehat{F}_j(x) \Big) \bigg)_{j=1, \ldots, d} $$ for all $x = (x^{1}, \ldots, x^{d}) \in \rset^d$ where $\widehat{F}_j (x) \defeq \frac{1}{n+1}\sum_{i=1}^n \mathds{1}\{X_i^j \leq x\}$ is the $j^{th}$ empirical marginal distribution.
Denoting by $V_i$ the standardized variables, $\forall i \in \{1, \ldots, n \}, V_i = \hat{T}(X_i)$.  The marginal distributions of $V_i$ are well approximated by  standard Pareto distribution, the approximation error comes from the fact that the empirical \emph{c.d.f}'s are used in $\widehat T$ instead of the genuine marginal \emph{c.d.f.}'s $F_j$. After this standardization step, the selected extreme samples are  $\{ V_i, \|V_i\| \geq  V_{(\lfloor \kappa n \rfloor)} \}$. 
\subsection{Enforcing regularity assumptions in Theorem~\ref{thm:main}}\label{sec:unifCV_app}
The methodology in the  present paper consists in learning a representation $Z$ for text data \emph{via} \HTalgo\ satisfying the regular variation  condition~(\ref{RV}). This condition   is weaker than the assumptions from Theorem~\ref{thm:main} for two reasons: first, it does not imply that  each class (conditionally to the label $Y$) is regularly varying, only that the distribution of $Z$  (unconditionally to the label) is. Second, 
in \citet{jalalzai2018binary}, 
it is additionally required that the regression function $\eta(z) =\PP{Y = +1 \given Z=z}$ converges uniformly as $\|z\|\to \infty$. Getting into details, one needs to introduce a limit random pair $(Z_\infty, Y_\infty)$ which distribution is the limit of $\PP{Y = \,\cdot\, ,  t^{-1} Z \in \, \cdot\, \given \|Z\| > t}$ as $t\to \infty$. Denote by $\eta_\infty$ the limiting regression function, $\eta_\infty(z)=  \PP{Y_\infty = +1 \given Z_\infty = z}$. The required assumption is that 
\begin{equation}
    \label{eq:uniformCV_regression}
\sup_{\{z \in \rset_+^d: \|z\| >t\}} \big| \eta(z) - \eta_\infty(z)\big| \xrightarrow[t\to\infty]{} 0. 
\end{equation}
 Uniform convergence~(\ref{eq:uniformCV_regression}) is not enforced in \HTalgo\, and the question of how to enforce it together with regular variation of each class separately remains open.  However, our experiments in sections~\ref{sec:Exp} and~\ref{sec:Exp_g} demonstrate  that enforcing  Condition~(\ref{RV})  is enough for our purposes,  namely improved classification and label preserving data augmentation.

\subsection{Logistic distribution} 
\label{logistic_Appendix}
\begin{figure}[tbh]
\begin{subfigure}[t]{0.32\textwidth}
    \includegraphics[width=\textwidth]{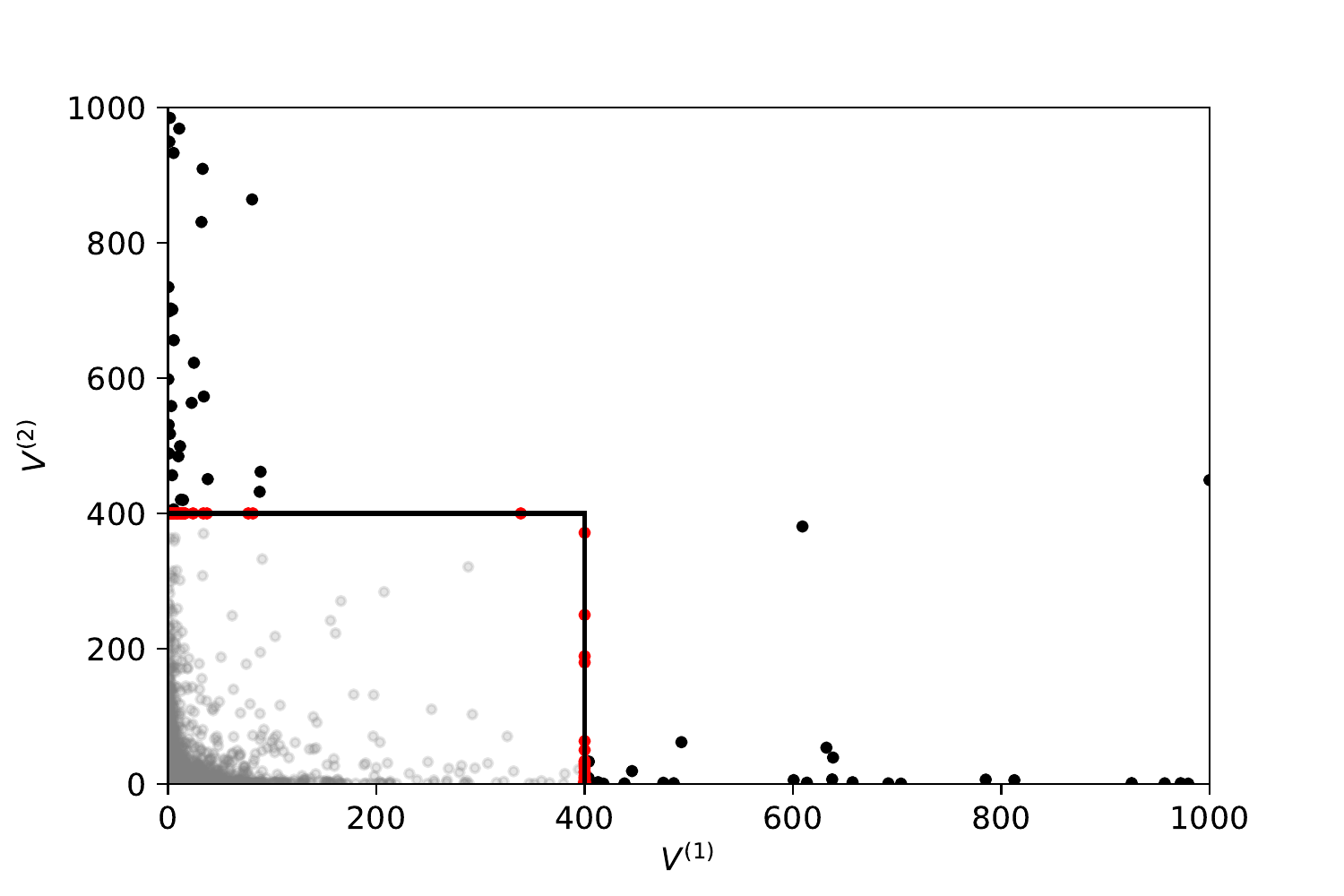}
    \caption{ near tail independence}
    \label{fig:logistic_alpha_09}
    \end{subfigure}
\begin{subfigure}[t]{0.32\textwidth}
    \includegraphics[width=\textwidth]{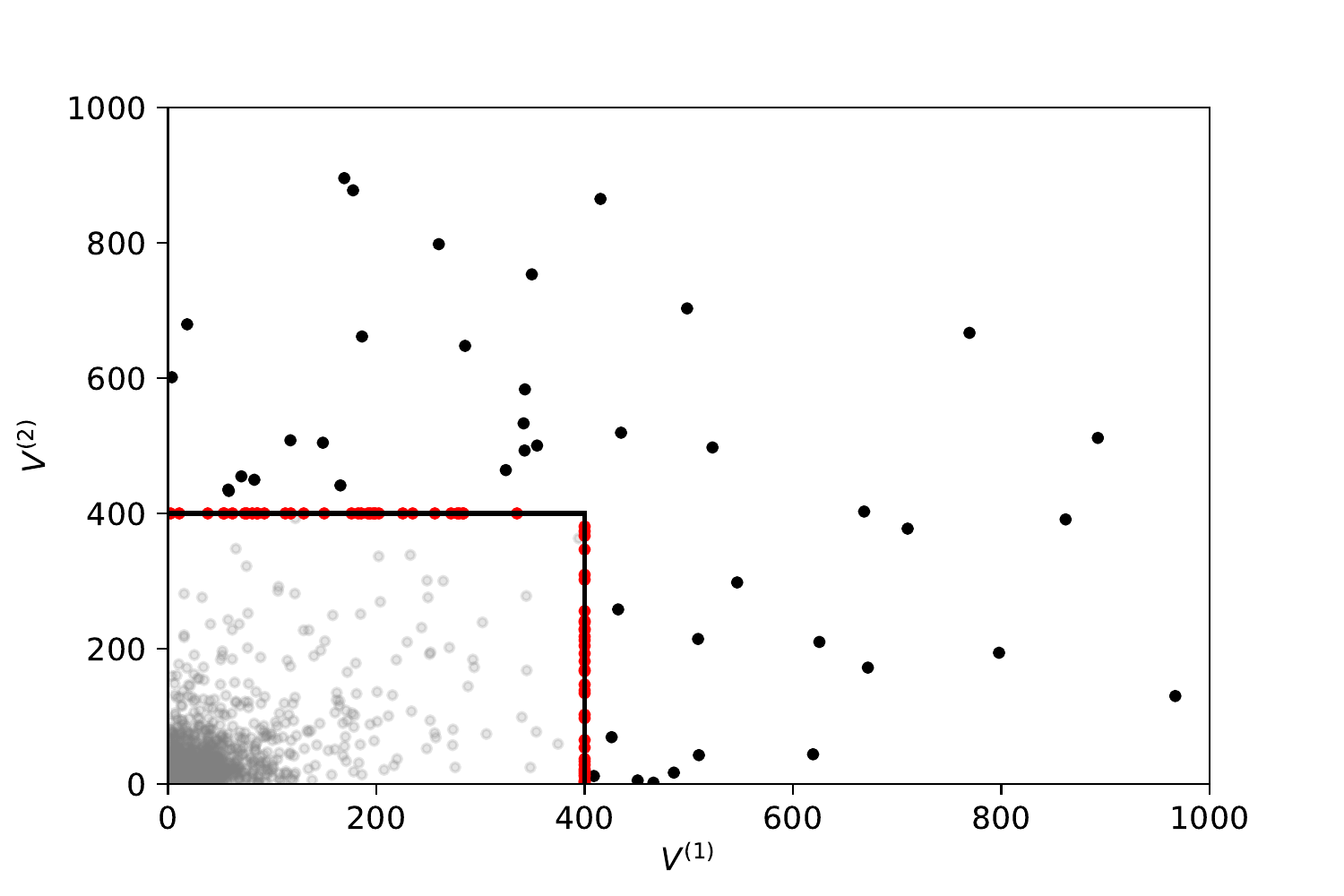}
    \caption{ moderate tail dependence}
    \label{fig:logistic_alpha_05}
        \end{subfigure}
\begin{subfigure}[t]{0.32\textwidth}
    \includegraphics[width=\textwidth]{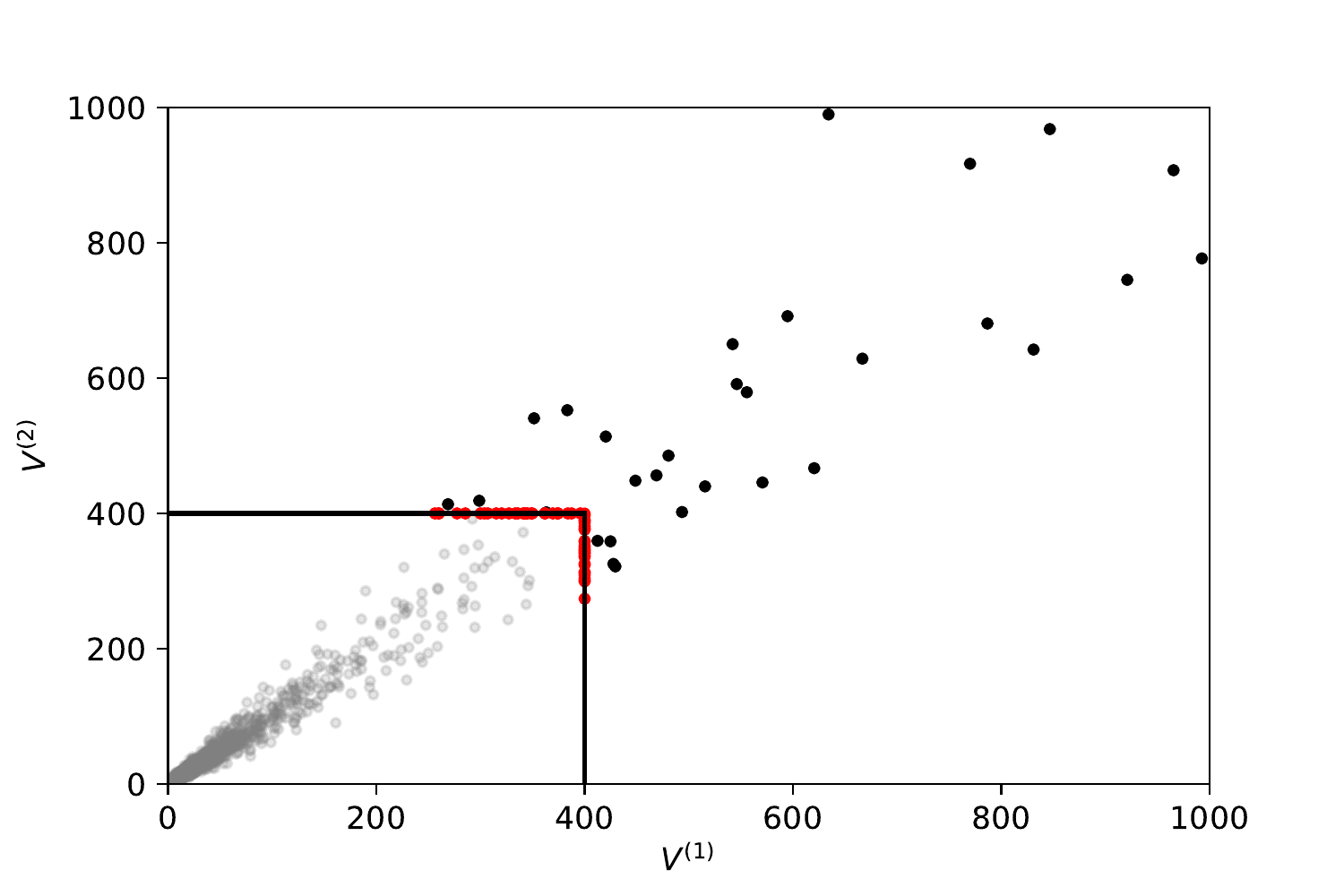}
    \caption{high tail dependence}
    \label{fig:logistic_alpha_01}
\end{subfigure}
\caption{Illustration of the distribution of the angle $\Theta(X)$ obtained with bivariate samples $X$ generated from a logistic model with different coefficients of dependence ranging from near asymptotic independence  \autoref{fig:logistic_alpha_09} ($\delta = 0.9$) to high asymptotic dependence \autoref{fig:logistic_alpha_01} ($\delta = 0.1$) including moderate dependence \autoref{fig:logistic_alpha_05} ($\delta = 0.5$). Non extreme samples are plotted in gray, extreme samples are plotted in black and the angles  $\Theta(X)$ (extreme samples projected on the sup norm sphere) are plotted in red. Note that not all extremes are shown since the plot was truncated for a better visualization. However all projections on the sphere are shown.}
\label{fig:Logistic_Exemples}
\end{figure}

 The logistic distribution  with dependence  parameter $\delta \in (0, 1]$  is defined in $\rset^d$ by its  \textit{c.d.f.}
$F(x) = \exp\big\{ - (\sum_{j=1}^d {x^{(j)}}^\frac{1}{\delta})^{\delta} \big\}$.
Samples from the  logistic distribution can be  simulated according to the algorithm proposed in  \citet{stephenson2003simulating}. \autoref{fig:Logistic_Exemples} illustrates this distribution with various values of $\delta$. Values of $\delta$ close to~$1$ yield  non concomitant extremes, \emph{i.e.} the probability of a simultaneous excess of a high threshold by more than one vector component is negligible. Conversely, for small values of $\delta$, extreme values tend to occur simultaneously. These two distinct tail dependence structures are respectively called  `asymptotic independence' and `asymptotic dependence' in the EVT terminology.

\subsection{Scale invariance comparison of BERT and \HTalgo}
\label{BERT_notRV}
In this section, we compare \HTalgo\ and BERT and show that the latter is not scale invariant. For this preliminary experiment we rely on labeled fractions of both \textit{Amazon} and \textit{Yelp} datasets respectively denoted as \textit{Amazon small dataset} and \textit{Yelp small dataset} detailed in \cite{kotzias2015group}, each of them containing $1000$ sequences 
from the large dataset. Both datasets are divided at random in a train set $\mathcal{T}_\text{train}$ and  $\mathcal{T}_\text{test}$. The train set represents \nicefrac{3}{4} of the whole dataset while the remaining samples represent the test set. We use the hyperparameters reported in Table
~\ref{tab:small_dataset_experiment}. 

\begin{table}[ht]
    \centering
    \begin{tabular}{c|ccc}\hline
     & NN model & $ \HTalgo_{1}$  & \HTalgo  \\ \hline
    Sizes of the layers $\varphi$   &  [768,384,200,50,8,1] & [768,384,200,100] & [768,384,200,150]   \\
    Sizes of the layers  $C_{\theta^\prime}^{bulk}$  & X  &  [100,50,8,1] &  [150,75,8,1]  \\
        Sizes of the layers  $C_{\theta}^{ext}$ & X &  X &  [150,75,8,1]\\
  $\rho_3$   & X  & X & 0.001 \\ 
\end{tabular}
    \caption{Network architectures for \textit{Amazon small dataset} and \textit{Yelp small dataset }.  The weight decay is set to $10^5$, the learning rate is set to $5*10^{-4}$, the number of epochs is set to 500 and the batch size is set to $64$.}
    \label{tab:small_dataset_experiment}
\end{table}

\begin{figure}[h]
    \centering
    \includegraphics[width=0.5\textwidth]{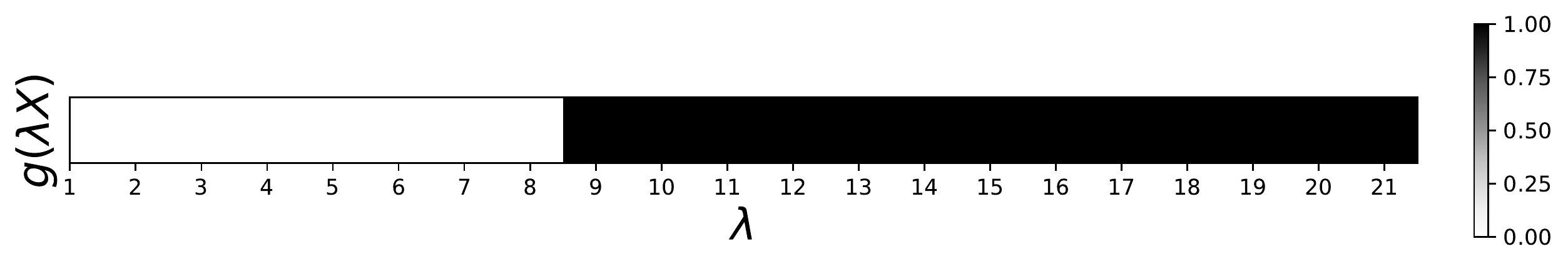}
    \caption{Lack of scale invariance of the classifier trained on BERT: evolution of the predicted label $g(\lambda X)$ from $-1$ to $+1$ for increasing values of $\lambda$, for one particular example $X$.}
    \label{fig:barcode_bert}
\end{figure}

\textbf{BERT is not regularly varying.}
In order to show that $X$ is not regularly varying, independence between $\|X\|$ and a margin of $\Theta(X)$ can be tested   \cite{coles1994statistical}, which is easily  done \emph{via} correlation tests.  
Pearson correlation tests were run on the extreme samples  of BERT and \HTalgo\ embeddings of $\textit{Amazon small dataset}$ and $\textit{Yelp small dataset}$. The statistical tests were performed  between all margins of $\big(\Theta(X_i)\big)_{1 \geq i \geq n}$ and $\big(\|X_i\|\big)_{1 \geq i \geq n}$.  
\begin{figure}[H]
\centering
\begin{subfigure}[t]{0.45\textwidth}
    \includegraphics[width=\textwidth]{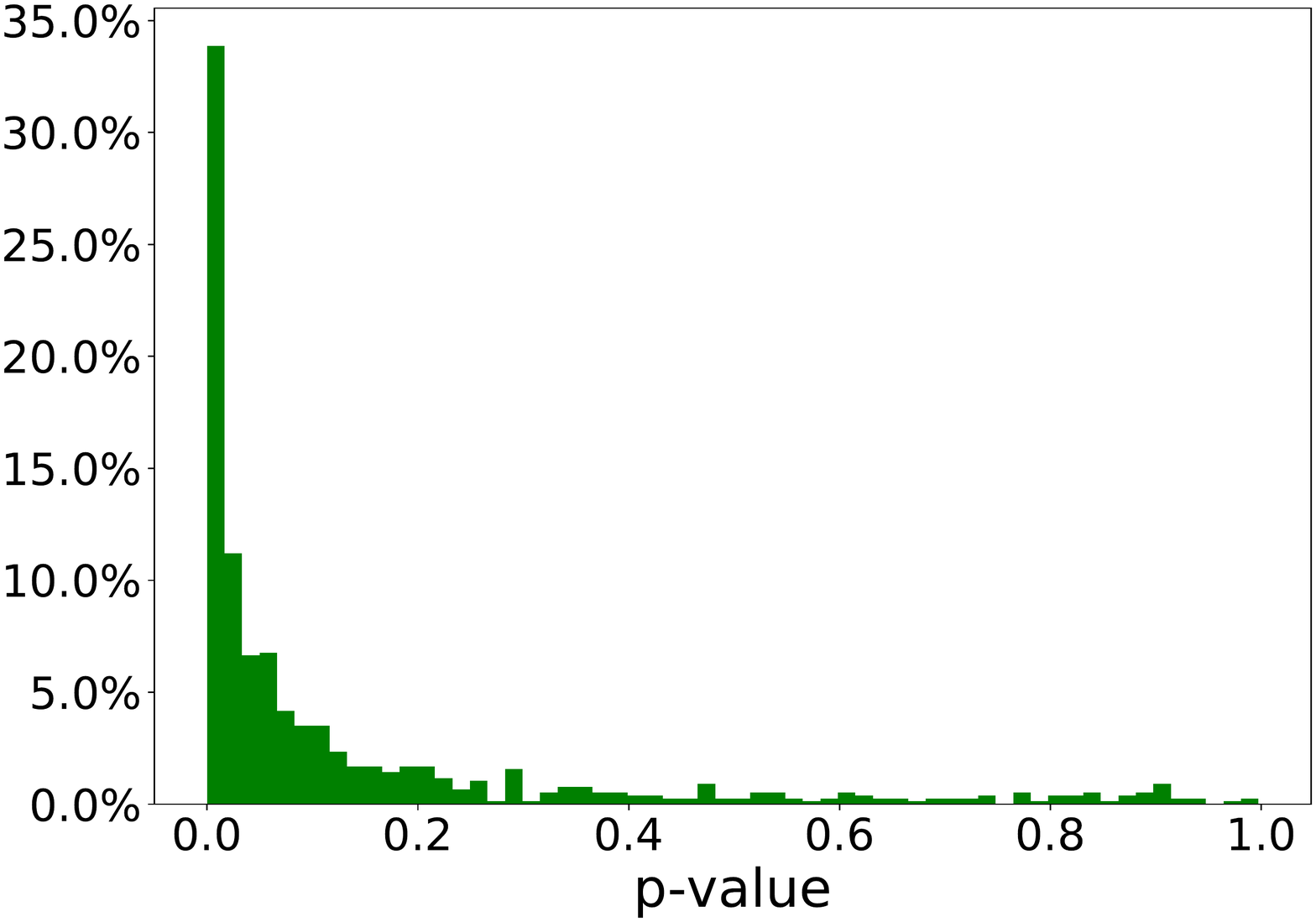}
    \caption{\textit{Yelp small dataset} - BERT}
    \label{fig:pearson_bert_yelp}
\end{subfigure}
\begin{subfigure}[t]{0.45\textwidth}
    \includegraphics[width=\textwidth]{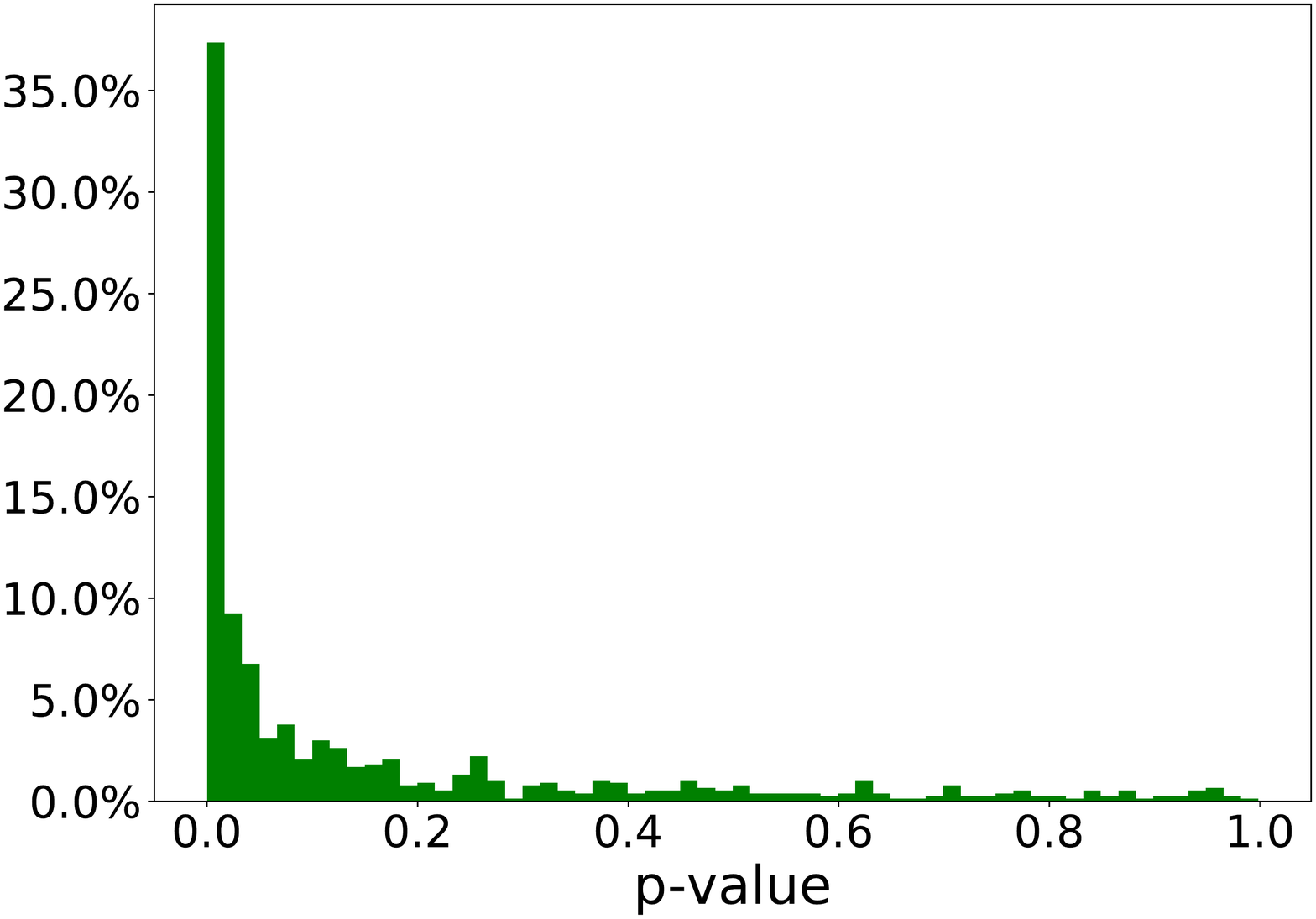}
    \caption{\textit{Amazon small dataset} - BERT}
    \label{fig:pearson_bert_amazon}
\end{subfigure}

\begin{subfigure}[t]{0.45\textwidth}
    \includegraphics[width=\textwidth]{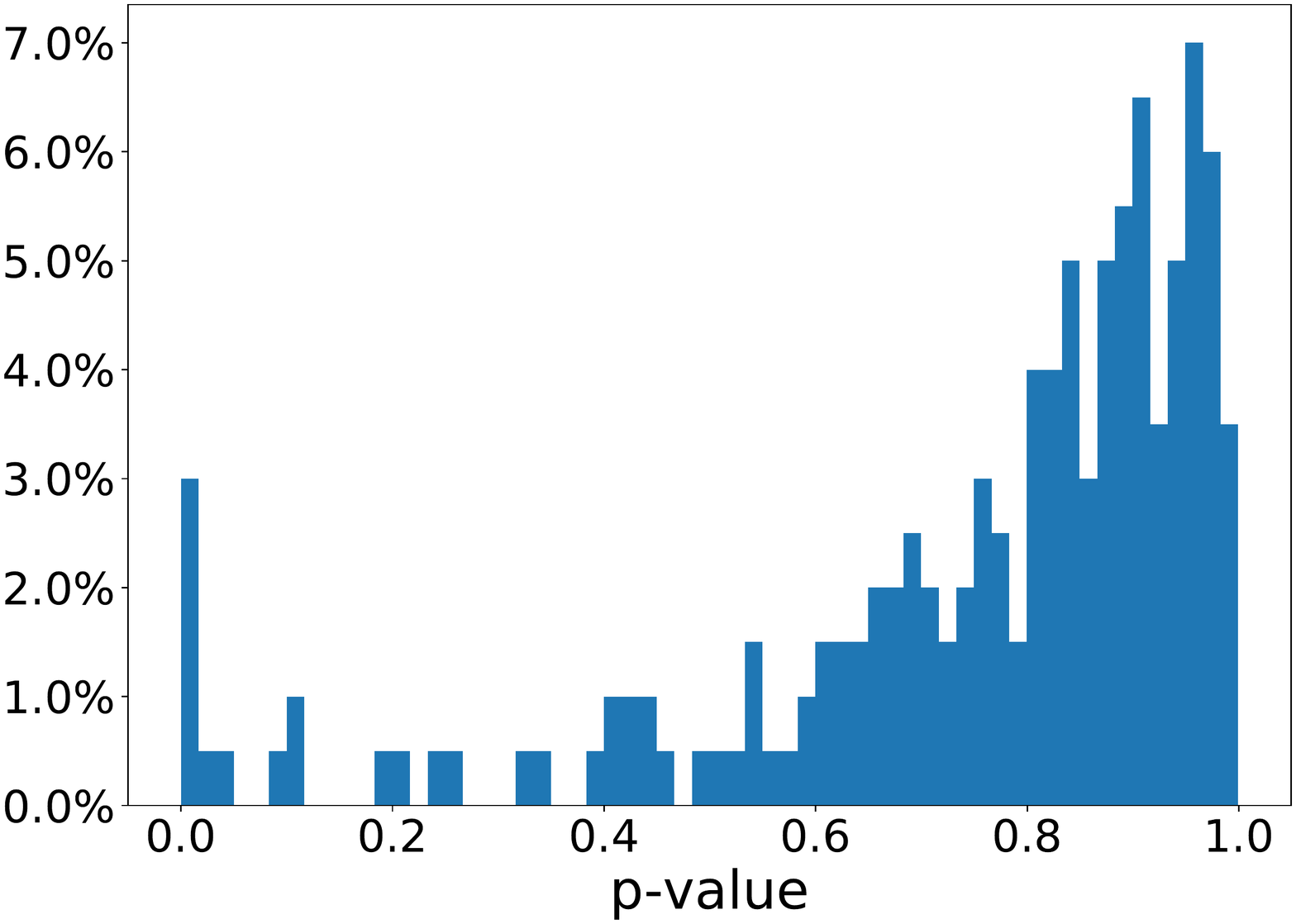}
    \caption{\textit{Yelp small dataset} - \HTalgo}
    \label{fig:pearson_lhtr_yelp}
\end{subfigure}
\begin{subfigure}[t]{0.45\textwidth}
    \includegraphics[width=\textwidth]{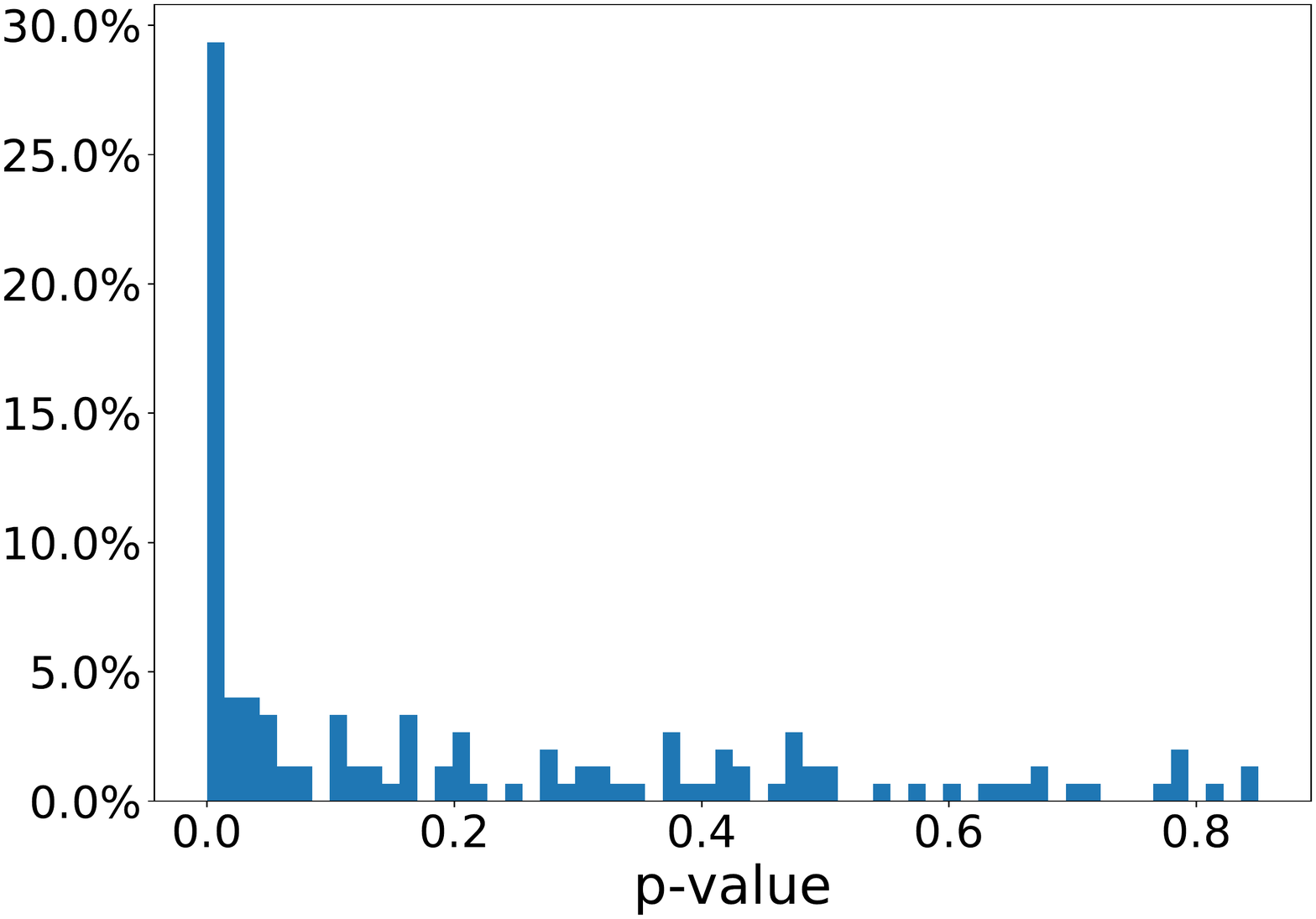}
    \caption{\textit{Amazon small dataset} - \HTalgo}
    \label{fig:pearson_lhtr_amazon}
\end{subfigure}
\caption{Histograms of the $p$-values for the non-correlation test between $\big(\Theta(X_i)\big)_{1 \geq i \geq n}$ and $\big(\|X_i\|\big)_{1 \geq i \geq n}$ on embeddings provided by BERT (Figure~\ref{fig:pearson_bert_yelp} and Figure~\ref{fig:pearson_bert_amazon}) or \HTalgo\ (Figure~\ref{fig:pearson_lhtr_yelp} and Figure~\ref{fig:pearson_lhtr_amazon}).}
\label{fig:pearson}
\end{figure}
Each histogram in Figure~\ref{fig:pearson}  displays the distribution of the  $p$-values of  the correlation tests between the margins  $X_j$ and the angle $\Theta(X)$ for $j \in \{1,\ldots d\}$, in  a given representation (BERT or \HTalgo) for a given dataset. 
For both \textit{Amazon small dataset} and \textit{Yelp small dataset} the distribution of the $p$-values is shifted towards larger values in the representation of \HTalgo\ than in BERT, which means that the correlations are weaker in the former representation than in the latter. This phenomenon is more pronounced with \textit{Yelp small dataset} than with \textit{Amazon small dataset}. Thus, in BERT representation, even the largest data points exhibit a non negligible correlation between the radius and the angle and the regular variation condition does not seem to be satisfied. 
As a consequence, 
in a classification setup such as binary sentiment analysis  detailed in Section~\ref{sec:sentiment_analysisYelp}), classifiers trained on BERT embedding are not guaranteed to be scale invariant. 
In other words for a representation $X$ of a 
sequence $U$ with a given label $Y$, the predicted label $g(\lambda X)$ is not necessarily constant for varying values of $\lambda \geq 1$.
Figure~\ref{fig:barcode_bert} illustrates this fact on a particular example taken from \textit{Yelp small dataset}. 
The color (white or black respectively) indicates the predicted class (respectively $-1$ and $+1$). For values of $\lambda$ close to $1$, the predicted class is $-1$ but the prediction shifts to class $+1$  for larger values of $\lambda$. 
{}

\begin{figure}[ht]
\centering
\begin{subfigure}[t]{0.45\textwidth}
    \includegraphics[width=\textwidth]{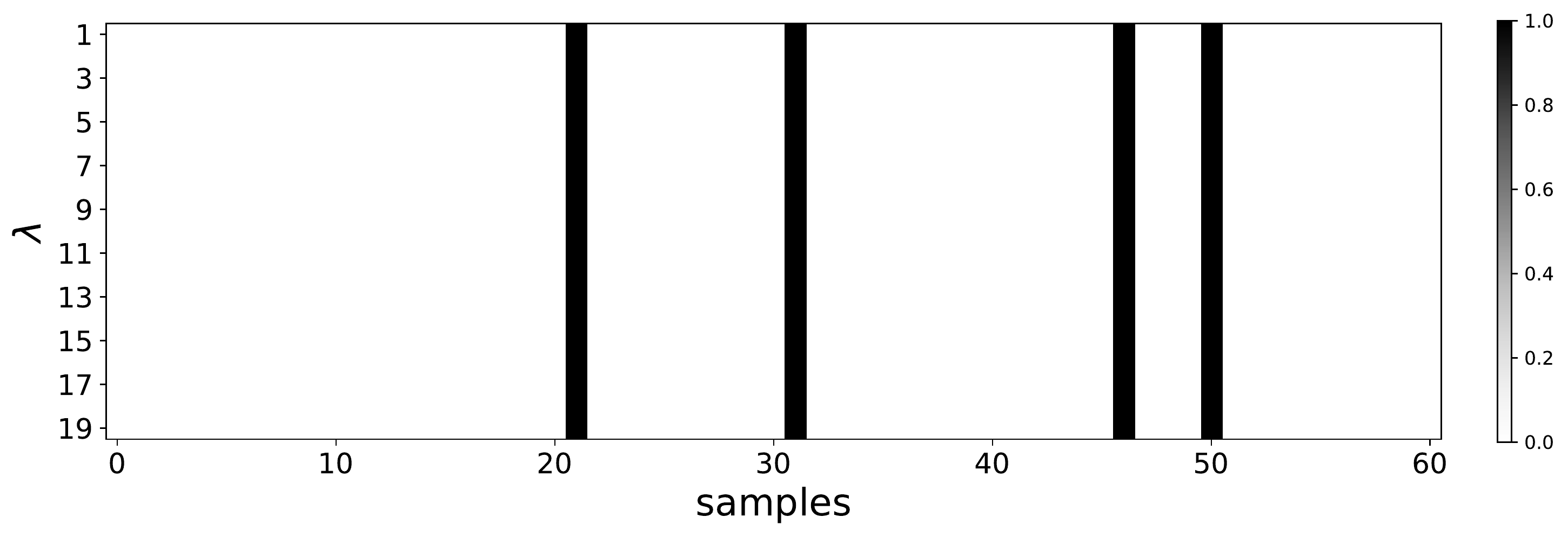}
    \caption{\textit{Amazon small dataset}}
    \label{fig:bar_code_amazon}
\end{subfigure}
\begin{subfigure}[t]{0.45\textwidth}
    \includegraphics[width=\textwidth]{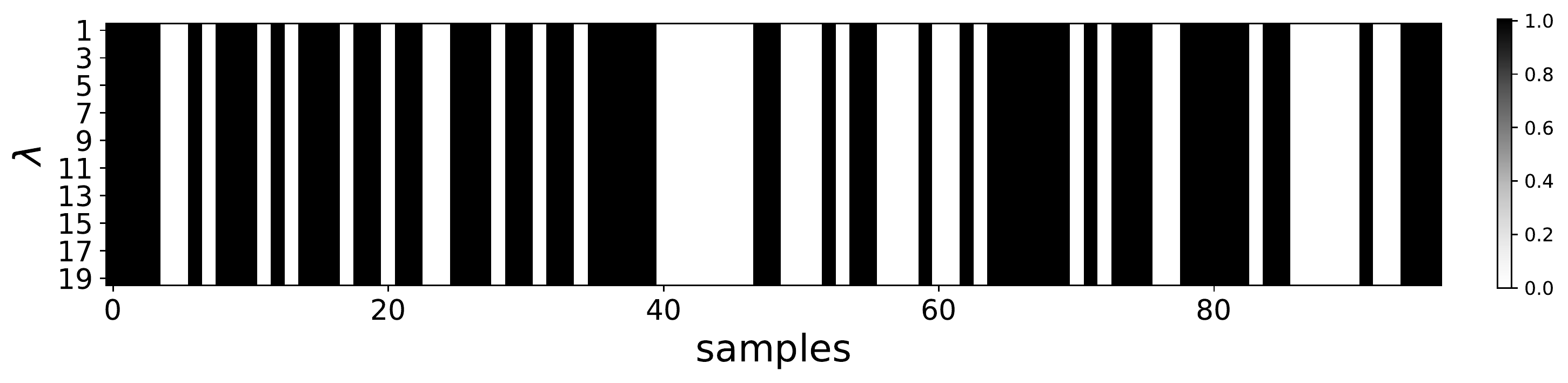}
    \caption{\textit{Yelp small dataset}}
    \label{fig:bar_code_yelp}
\end{subfigure}
\caption{Scale invariance of $g^\text{ext}$ trained on LHTR: evolution of the predicted label   $g^\text{ext}(\lambda Z_i)$ (white or black for $-1 / +1$) for increasing values of $\lambda$, for samples $Z_i$ from the extreme test set $\mathcal{T}_\text{test}$ from \textit{Amazon small dataset} (Figure~\ref{fig:bar_code_amazon}) and  \textit{Yelp small dataset} (Figure~\ref{fig:bar_code_yelp})\label{fig:barcodes_lhtr}.}
\end{figure}

\textbf{Scale invariance of \HTalgo.} 
We provide here experimental evidence that \HTalgo's classifier $g^\text{ext}$ is scale invariant (as defined in Equation~(\ref{eq:invariance_dilation})). Figure~\ref{fig:barcodes_lhtr} displays the predictions $g^{\text{ext}}(\lambda Z_i)$ for increasing values of the scale factor $\lambda\ge 1$  and $Z_i$ belonging to $\mathcal{T}_{\text{test}}$, the set of samples considered as extreme in the learnt representation. For any such sample $Z$,
the predicted label remains constant as $\lambda$ varies, \emph{i.e.} it is  scale invariant, $g^\text{ext}(\lambda Z) = g^\text{ext}(Z)$, for all $\lambda\ge 1$. 

\subsection{Experimental settings (Classification): additional details}
\label{additional_exp_classif}

\textbf{Toy example. } For the toy example, we generate $3000$ points distributed as a mixture of two normal distributions in dimension two. 
For training \HTalgo, the number of epochs is set to $100$ with a  dropout rate equal to $0.4$, a  batch size of $64$ and a learning rate of $5*10^{-4}$. The weight parameter $\rho_3$ in the loss function (Jensen-Shannon divergence from the target) is set to $10^{-3}$. 
Each component $\varphi$, $C^{\text{bulk}}$ and $C^{\text{ext}} $ is made of $3$ fully connected layers, the sizes of which are reported in  Table~\ref{tab:toys_dataset_experiment}. \\
\textbf{Datasets.} For Amazon, we
  work with the video games subdataset from
  \url{http://jmcauley.ucsd.edu/data/amazon/}. For Yelp \cite{yelp_1,yelp_2}, we work with 1,450,000 reviews after that can be found at
  \url{https://www.yelp.com/dataset}. 


\begin{table}[ht]
    \centering
    \begin{tabular}{c|c}\hline
&     Layers' sizes \\ \hline
    $\varphi$  
    & [2,4,2]   \\
    $C_{\theta^\prime}^{bulk}$ 
    &  [2,8,1]   \\
        $C_{\theta}^{ext}$ 
        &  [2,8,1] \\
\end{tabular}
    \caption{ Sizes of the successive layers in each component of \HTalgo\ used in the toy example. }
    \label{tab:toys_dataset_experiment}
\end{table}

\textbf{BERT representation for text data.} 
We use BERT pretrained models and code from the library \textit{Transformers} \footnote{\url{https://github.com/huggingface/transformers}}. All models were implemented using Pytorch and trained on a single Nvidia P100. The output of BERT is a $\rset^{768}$ vector. All parameters of the models have been selected using the same grid search. 

\textbf{Network architectures.}
Tables~\ref{tab:huge_dataset_experiment} report the architectures (layers sizes) chosen for each component of the three algorithms considered for performance comparison (Section~\ref{sec:Exp}), respectively for the moderate and large datasets used in our experiments.  We set $\rho_1 = (1 -  \hat{\mathbb{P}}(||Z|| \geq ||Z_{(\lfloor \kappa n \rfloor)}||))^{-1}$ and $\rho_2=\hat{\mathbb{P}}(||Z||\geq ||Z_{(\lfloor \kappa n \rfloor)}||)^{-1}$.


\begin{table}[ht]
    \centering 
    \begin{tabular}{c|ccc}\hline
     & NN model & $ \HTalgo_{1}$  & $ \HTalgo$    \\\hline
   Sizes of the layers $\varphi$   &  [768,384,200,50,8,1] & [768,384,200,100] & [768,384,200,150]   \\
    Sizes of the layers of  $C_{\theta^\prime}^{bulk}$ & [150,75,8,1]  &  [100,50,8,1] &  [150,75,8,1]  \\
        Sizes of the layers of  $C_{\theta}^{ext}$ & X &  X &  [150,75,8,1]\\
  $\rho_3$   & X  & X & 0.01 \\ 
\end{tabular}
    \caption{Network architectures for \textit{Amazon dataset} and \textit{Yelp dataset}. The weight decay is set to $10^5$, the learning rate is set to $1*10^{-4}$, the number of epochs is set to 500 and the batch size is set to $256$.}
    \label{tab:huge_dataset_experiment}
\end{table}

\subsection{Experiments for data generation}
\label{additional_exp_generation}
\subsubsection{Experimental setting}
As mentioned in Section~\ref{subsection:generation}, hyperparameters for dataset augmentation are detailed in Table~\ref{tab:generation_experiments}.
\begin{table}[ht]
    \centering
    \begin{tabular}{c|ccc}\hline
     &  \HTalgo  \\\hline 
   Sizes of the layers $\varphi$  & [768,384,200,150]   \\ 
    Sizes of the layers of  $C_{\theta^\prime}^{bulk}$ & [150,75,8,1]  \\ 
        Sizes of the layers of  $C_{\theta}^{ext}$ & [150,75,8,1]\\ 
  $\rho_3$ &0.01 \\ 
\end{tabular}
    \caption{ For \textit{Amazon}  and \textit{Yelp}, we follow \cite{classif} the weight decay is set to $10^5$, the learning rate is set to $1*10^{-4}$, the number of epochs is set to 100 and the batch size is set to $256$.}
    \label{tab:generation_experiments}
\end{table}
For the Transformer Decoder we use $2$ layers with $8$ heads, the dimension of the key and value is set to $64$ \cite{attention_is} 
and  the inner dimension is set to $512$. The architectures for the models proposed by \citet{eda} and \citet{naacl_baseline} are chosen according to the original papers. For a fair comparison with \citet{naacl_baseline}, we update the language model with a BERT model, the labels are embedded in $\rset^{10}$ and  fed to a single MLP layer (this dimension is chosen using the same procedure as in \cite{colombo2019affect,dinkar2020importance}). The new model is trained using AdamW \cite{adamW}.
\subsubsection{Influence of the scaling factor on the linguistic content } 
Table~\ref{tab:additional_example_sentences} gathers some extreme sequences 
generated by \geneliex\ for $\lambda$ ranging from $1$ to $1.5$.  
No major linguistic change appears when $\lambda$ varies. The generated sequences  
are grammatically correct and share the same polarity (positive or negative sentiment) as the input sequence. 
Note that for greater values of $\lambda$, a repetition phenomenon appears. The resulting sequences 
keep the label and polarity of the input 
sequence but repeat some words \cite{holtzman2019curious}.
 \begin{table}[ht]
\centering
\resizebox{.9\textwidth}{!}{ \begin{tabular}[t]{c|p{16cm}}
\small Input	& very sloppy and slow service. when we arrived they told us to sit anywhere but all the tables were still dirty and haven't been cleaned. they didn't bother to ask if we wanted refills on our drinks. we needed an extra plate and didn't get one so my nephew decides to go up to the counter and ask for one because he's hungry. they gave our check when we were still eating. the list can go on and on. i wouldn't recommend this place. go somewhere else for faster and better service. very disappointed
\\
\small $\lambda = 1.1$	&very sloppy and sluggish service. when we got there, they told us to sit anywhere but all the tables were empyt full of dishes and were not cleaned at all. they didn't bother to ask if our drinks would be added. we needed an extra dish and didn't get one, so my cousin decided to go to the counter and ask one because he's hungry. they were going to watch while we were still eating. the list could go on and on. i would not recommend this place. go elsewhere for faster and better service. very very disappointed \\
\small $\lambda = 1.2$	&services and survivors. when he got there, he told us we were sitting everywhere but all the tables were full of dishes and we didn't wash everything. he never bothered to ask if our drinks would be added. we needed extra food and didn't get one, so my brother decided to go to the locker and ask because he was thirsty. they want to watch it while we eat. the list can be continuous and active. i would not recommend this place. go elsewhere for faster and better service. very disappointed \\
\small $\lambda = 1.3$	&services and survivors. when he got there, he told us that we were sitting everywhere, but all the tables were full of dishes and we didn't wash everything. he never bothered to ask if our drinks would be added. We needed more food and we didn't get it, so my brother decided to go to the locker and ask because he was thirsty. they want to watch it when we eat. the list can be continuous and active. i would not recommend this place. go faster and faster for better service. very disappointed \\\hline
\small Input	&visited today with my husband. we were in the firearms section. there were 3 employees in attendance with one customer. my husband ask a question and was ignored. he waited around for another 10 minutes or so. if it had been busy i could understand not receiving help. we left and went elsewhere for our purchases. \\
\small $\lambda = 1.1$	&visited today with my husband. we were in the firearms section. together with one customer there were 3 employees. my husband asked and was ignored. waited about another 10 minutes. if it was busy, i would understand that i wouldn't get help. we left and went somewhere else because of our purchases. \\
\small $\lambda = 1.2$	&today she visited with her husband. we were in the gun department. there were 3 employees together with one customer. my husband asked and was ignored. waited another 10 minutes. if he was busy, i would understand that i would not receive help. we went and went somewhere else because of our shopping. \\
\small $\lambda = 1.3$	&today, she went with her husband. we are in the gun department. there are 3 employees and one customer. my husband rejected me and ignored him. wait another minute. if he has a job at hand, i will understand that i will not get help. we went somewhere else because of our business. \\\hline
\small Input	&walked in on a friday and got right in. it was exactly what i expected for a thai massage. the man did a terrific job. he was very skilled, working on the parts of my body with the most tension and adjusting pressure as i needed throughout the massage. i walked out feeling fantastic and google eyed. \\
\small $\lambda = 1.1$	&walked in on a friday and got right in. it was exactly what i expected for a thai massage. the man did a terrific job. he was very skilled, working on the parts of my body with the most tension and adjusting pressure as needed throughout the massage. i walked out feeling fantastic and google eyed. \\
\small $\lambda = 1.2$	&climb up the stairs and get in. the event that i was expecting a thai massage. the man did a wonderful job. he was very skilled, dealing with a lot of stress and stress on my body parts. i walked out feeling lightly happy and tired. \\
\small $\lambda = 1.3$	&go up and up. this was the event i was expecting a thai massage. the man did a wonderful job. what this was was an expert, with a lot of stress and stress on my body parts. i walked out feeling lightly happy and tired. \\\hline
\small Input	&i came here four times during a 3 - day stay in madison. the first two was while i was working - from - home. this place is awesome to plug in, work away at a table, and enjoy a great variety of coffee. the other two times, i brought people who wanted good coffee, and this place delivered. awesome atmosphere. awesome awesome awesome. \\
\small $\lambda = 1.1$	&i came here four times during a 3-day stay in henderson. the first two were while i was working - from home. this place is great for hanging out, working at tables and enjoying the best variety of coffee. the other two times, i brought in people who wanted a good coffee, and it delivered a place. better environment. really awesome awesome. \\
\small $\lambda = 1.2$	&i came here four times during my 3 days in the city of henderson. the first two were while i was working - at home. this place is great for trying, working tables and enjoying the best variety of coffee. the other two times, i brought people who wanted good coffee, and it brought me somewhere. good environment. really amazing. \\
\small $\lambda = 1.3$	&i came here four times during my 3 days in the city of henderson. the first two are when i'm working - at home. this place is great for trying, working tables and enjoying a variety of the best coffees. the other two times, i bring people who want good coffee, and that brings me somewhere. good environment. very amazing.\\\hline
\end{tabular}}
\caption{
Sequences generated by \geneliex\ for extreme embeddings implying label (sentiment polarity) invariance for generated 
Sequence. $\lambda$ is the scale factor. Two first reviews are negatives, two last reviews are positive.}
\label{tab:additional_example_sentences}
\end{table}

\section{Extremes in Text}\label{explainabilite}

\paragraph{Aim of the experiments}  
The aim of this section is double:  first, to provide some intuition on what characterizes sequences falling in the extreme region of \HTalgo.  
Second, to investigate the hypothesis that extremes from \HTalgo\ are input sequences 
which tend to be harder to model than non extreme ones 

Regarding the first aim (\emph{ (i) Are there interpretable text features correlated with the extreme nature of a text sample?}, since we characterize extremes by their  norm in \HTalgo\ representation, in practice the question boils down to finding text features which are positively correlated with the norm of the text samples in \HTalgo, which we denote by $\|\varphi(X)\|$ and referred to as the `\HTalgo\ norm' in the sequel. 
Preliminary investigations did not reveal semantic features (related to the meaning or the  sentiment expressed in the sequence 
) displaying such correlation. However we have identified two  features which are positively correlated both together and with the norm in \HTalgo, namely the sequence length $|U|$ as measured by the  number of tokens of the input (recall that in our case an input sequence $U$ is a review composed of multiple sequences 
),  and the  norm of the input  in BERT representation (`BERT norm', denoted by $\|X\|$).  

As for the second question (\emph{ (ii) Are  \HTalgo's extremes harder to model?} ) we consider the next token prediction loss \cite{lm} (`LM loss' in the sequel) obtained by training a language model on top of BERT. 
The next token prediction loss can be seen as a measure of hardness to model the input sequence. 
The question is thus to determine whether this prediction loss is correlated with the  norm in \HTalgo\ (or in BERT, or with the sequence length).

\begin{figure}
    \centering
    
    \includegraphics[width=0.8\linewidth]{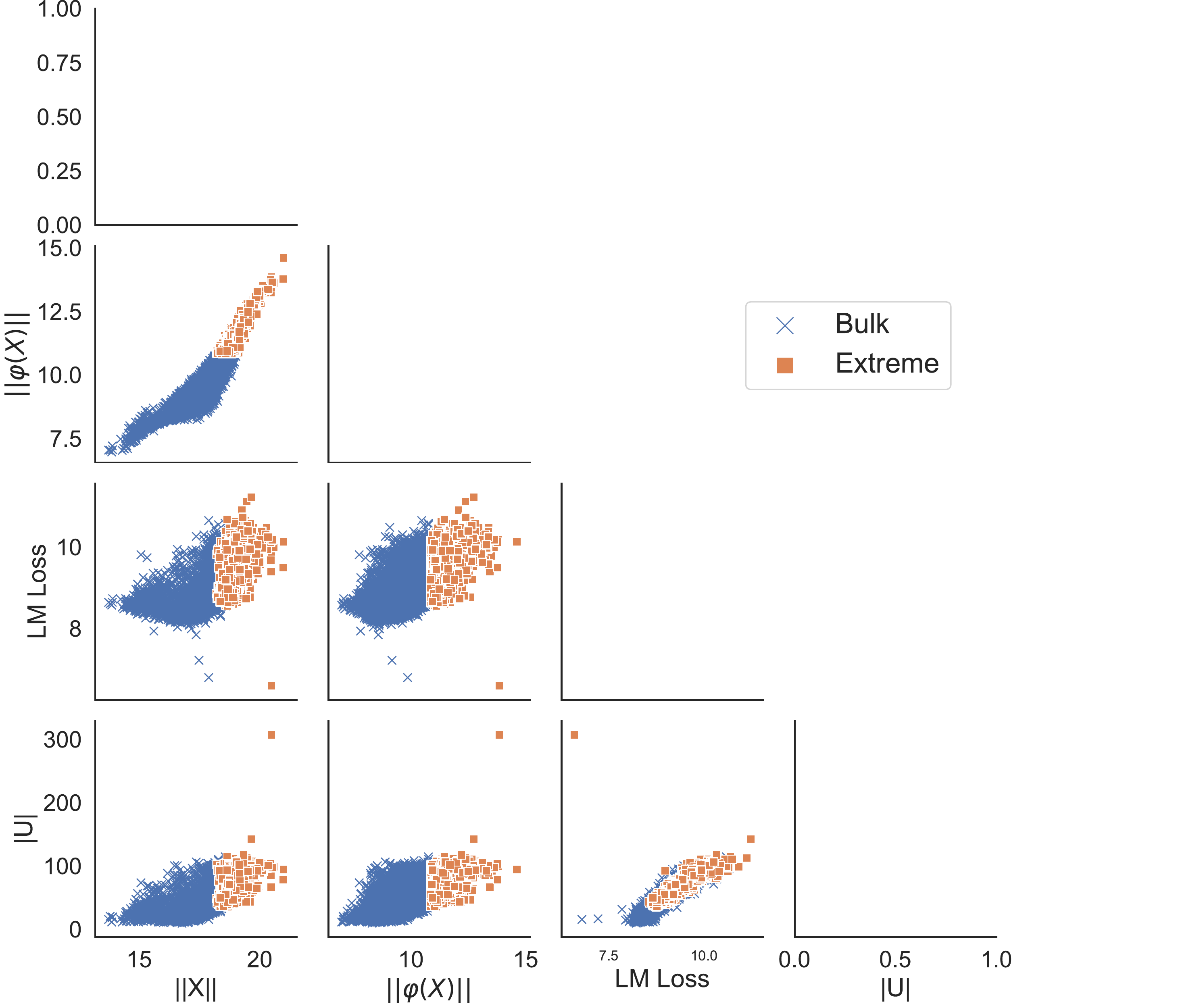} \\
   
    \includegraphics[width=0.8\linewidth]{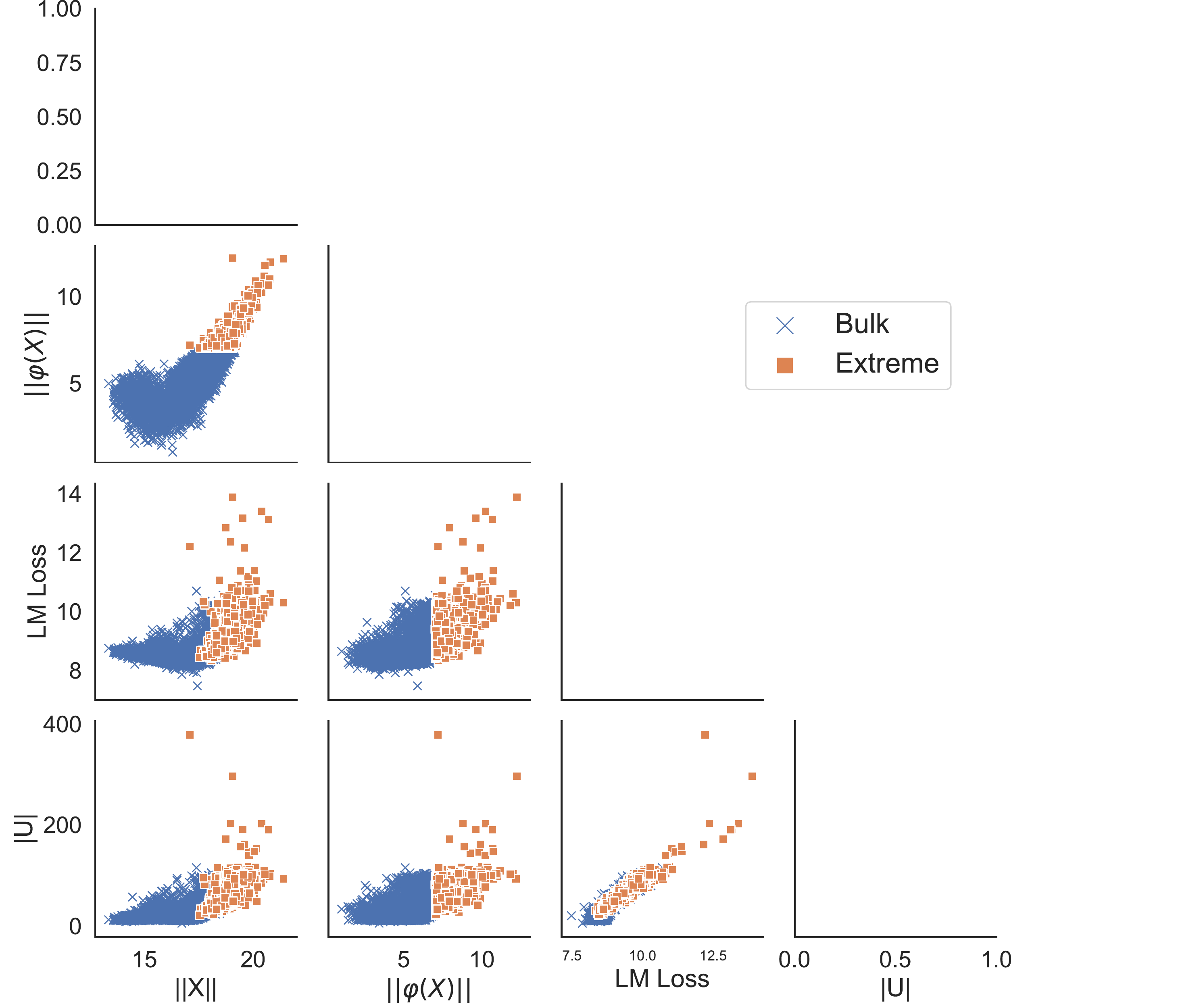}
    \caption{Scatterplots of the four variables  `BERT norm', `\HTalgo\ norm',  `LM loss' and `sequence length'  on \textit{Yelp} dataset (top) and \textit{Amazon} dataset (bottom).}
    \label{fig:pairplot_test_4variables}
\end{figure}

\begin{figure}
    \centering
    \includegraphics[width=0.45\linewidth]{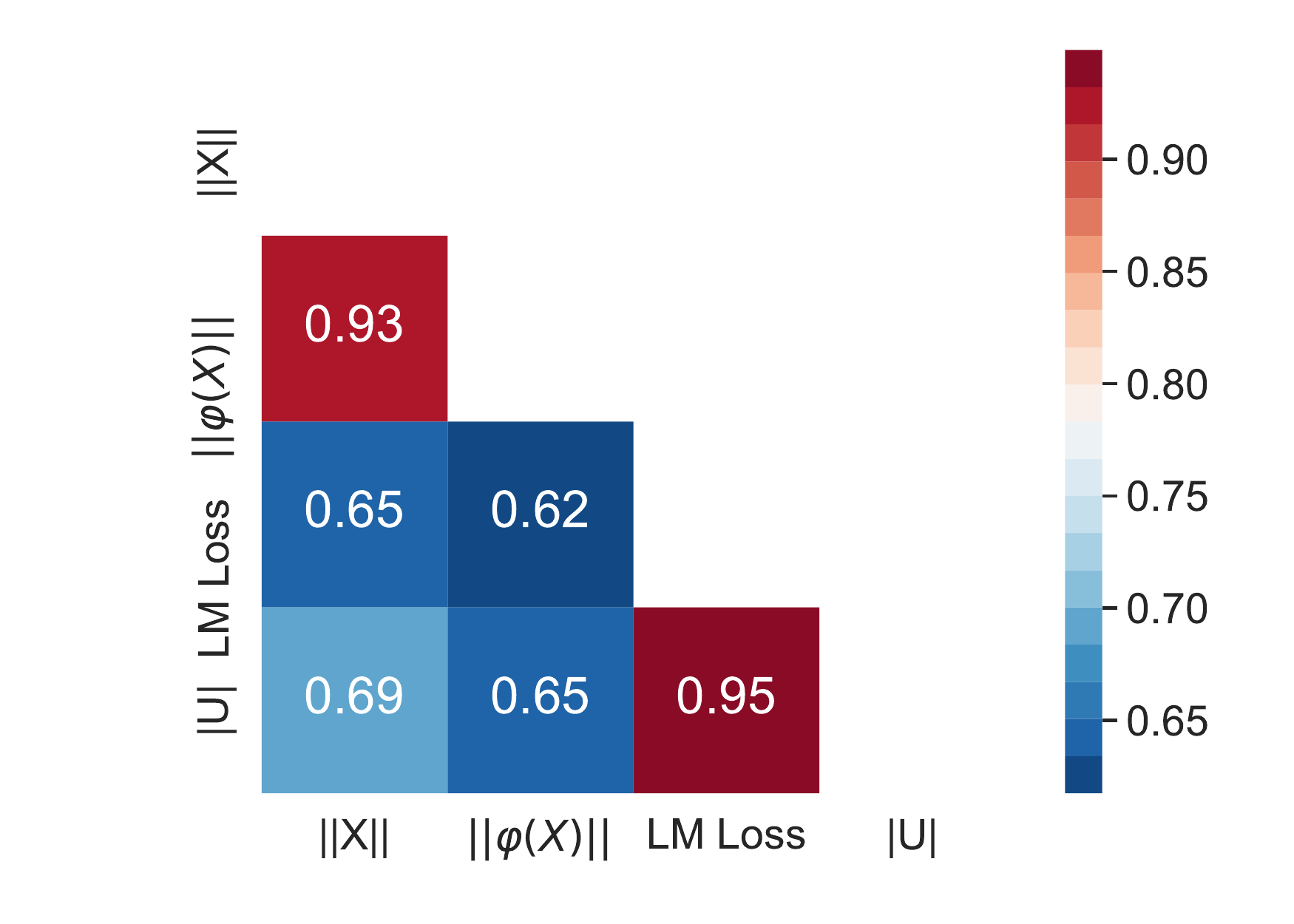}
    \includegraphics[width=0.45\linewidth]{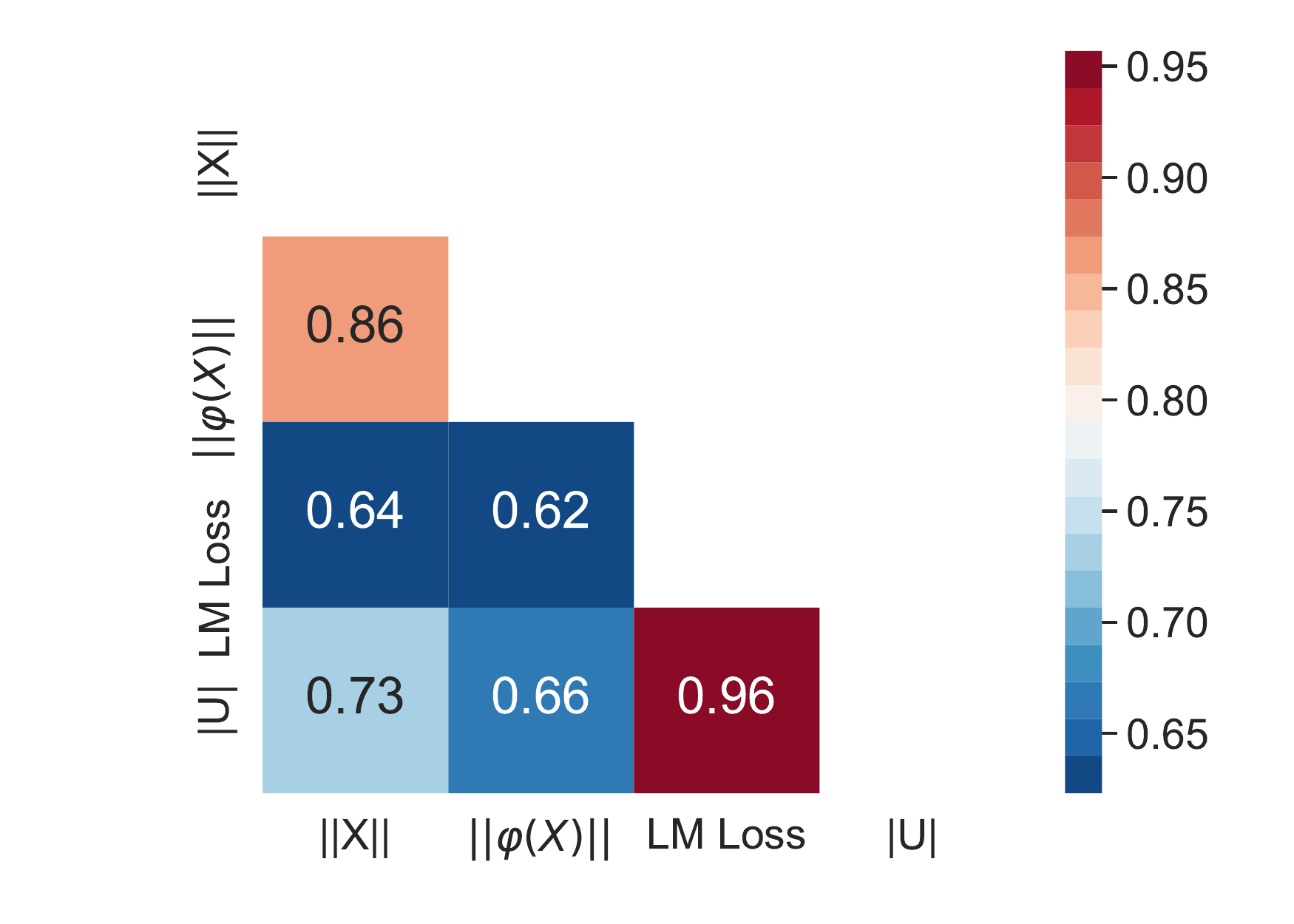}
    \caption{Non diagonal entries of the correlation matrices of the four variables  `BERT norm', `\HTalgo\ norm',  `LM loss' and `sequence length' for  \textit{Yelp} dataset (left) and \textit{Amazon} dataset (right). } 
    \label{fig:correl_test_4variables}
\end{figure}

\paragraph{Results}
 Figure~\ref{fig:pairplot_test_4variables} displays pairwise scatterplots  for the four considered variables  
on \textit{Yelp} dataset (left) and \textit{Amazon} dataset (right). These scatterplot suggest strong dependence for all pairs of variables. For a more quantitative assessment, 
Figure~\ref{fig:correl_test_4variables} displays  the correlation matrices between the four quantities   $\|\varphi(X)\|$, $\|X\|$, $|U|$ and `LM Loss' described above 
on Amazon and Yelp datasets. 
Pearson and Spearman two-sided correlation tests are performed on all pairs of variables, both tests having as null hypothesis that the correlation between two variables is zero. For all tests, $p$-values are smaller than $10^{-16}$, therefore null hypotheses are rejected for all pairs.

These results prove that the four considered variables are indeed significantly positively correlated, which answers questions $(i)$ and $(ii)$ above.  

\begin{figure}[H]
\centering
\begin{subfigure}[t]{0.45\textwidth}
    \includegraphics[width=\textwidth]{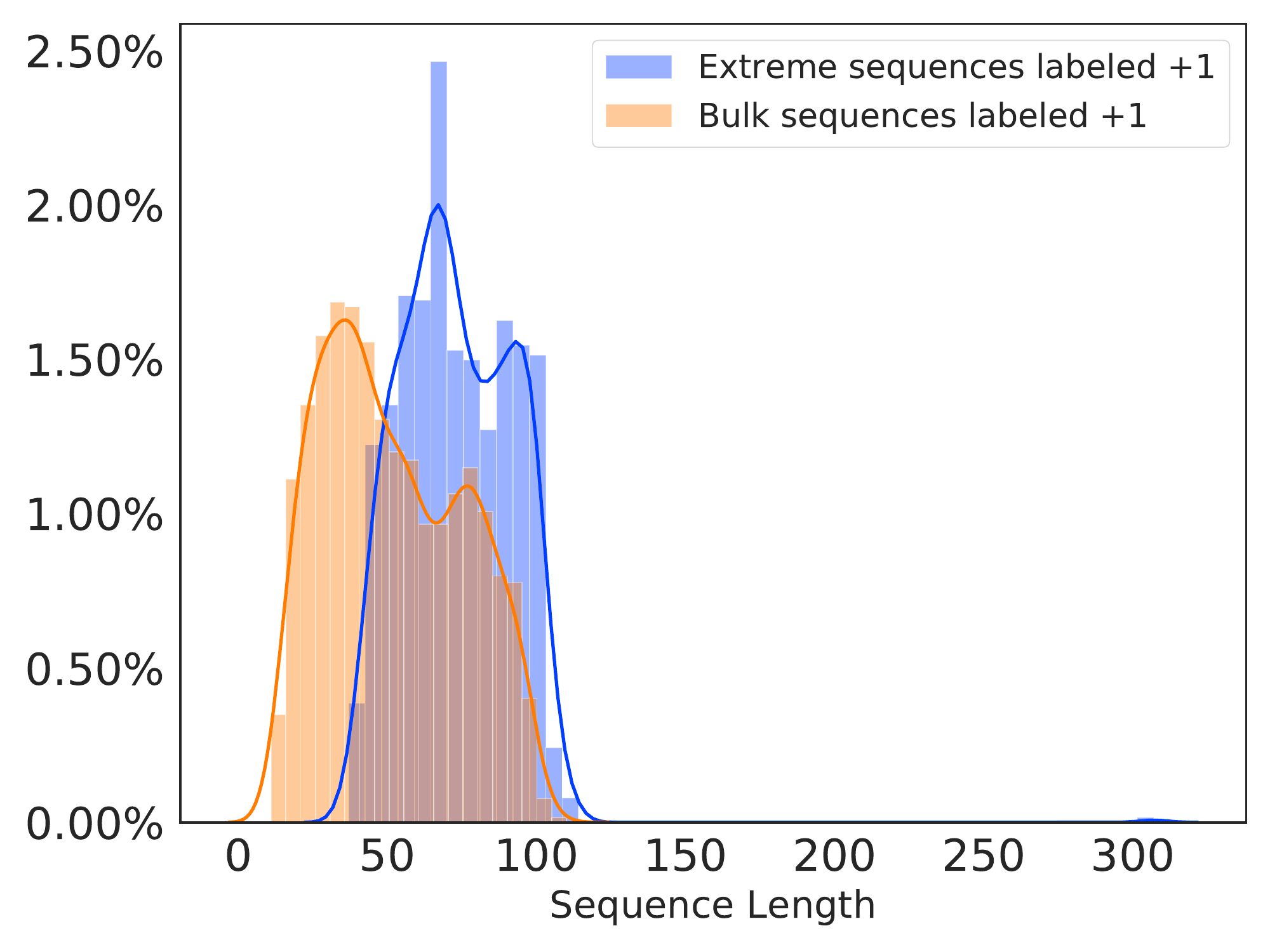}
    \caption{\textit{Yelp} - labeled $+1$ }
    \label{fig:Yelp+1}
\end{subfigure}
\begin{subfigure}[t]{0.45\textwidth}
    \includegraphics[width=\textwidth]{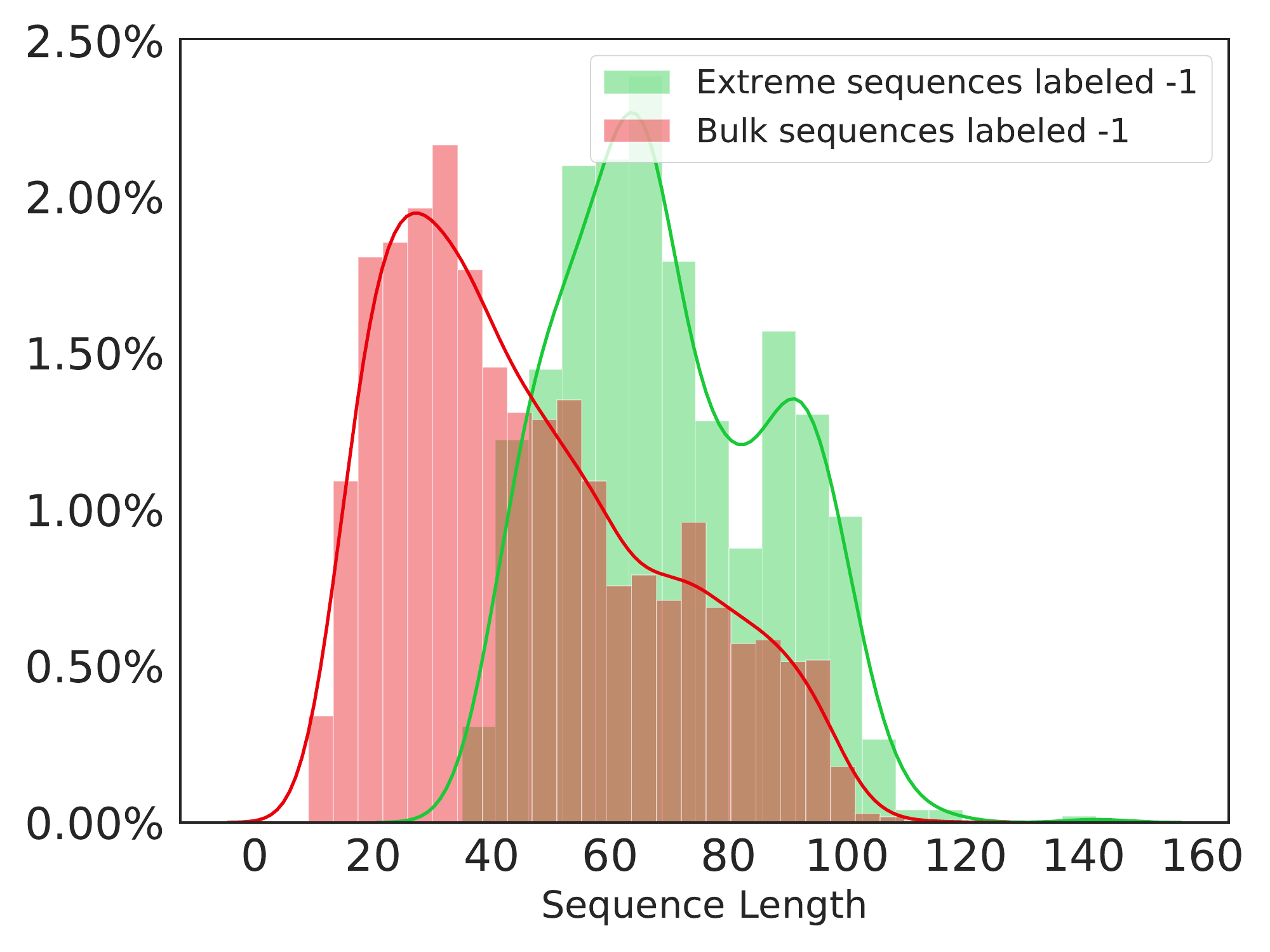}
    \caption{\textit{Yelp} -  labeled $-1$ }
    \label{fig:Yelp-1}
\end{subfigure}

\begin{subfigure}[t]{0.45\textwidth}
    \includegraphics[width=\textwidth]{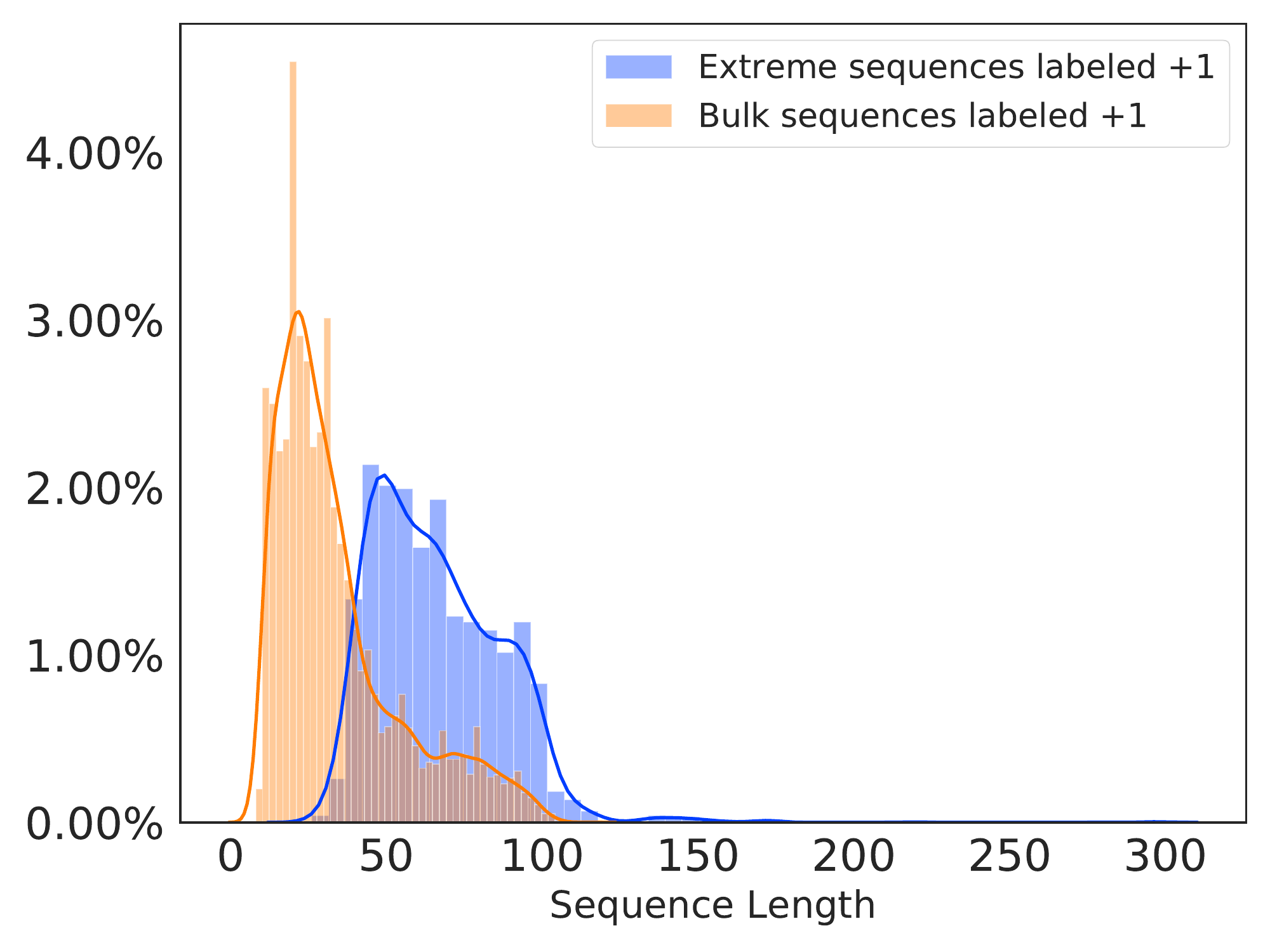}
    \caption{\textit{Amazon} -  labeled $+1$ }
    \label{fig:Amazon+1}
\end{subfigure}
\begin{subfigure}[t]{0.45\textwidth}
    \includegraphics[width=\textwidth]{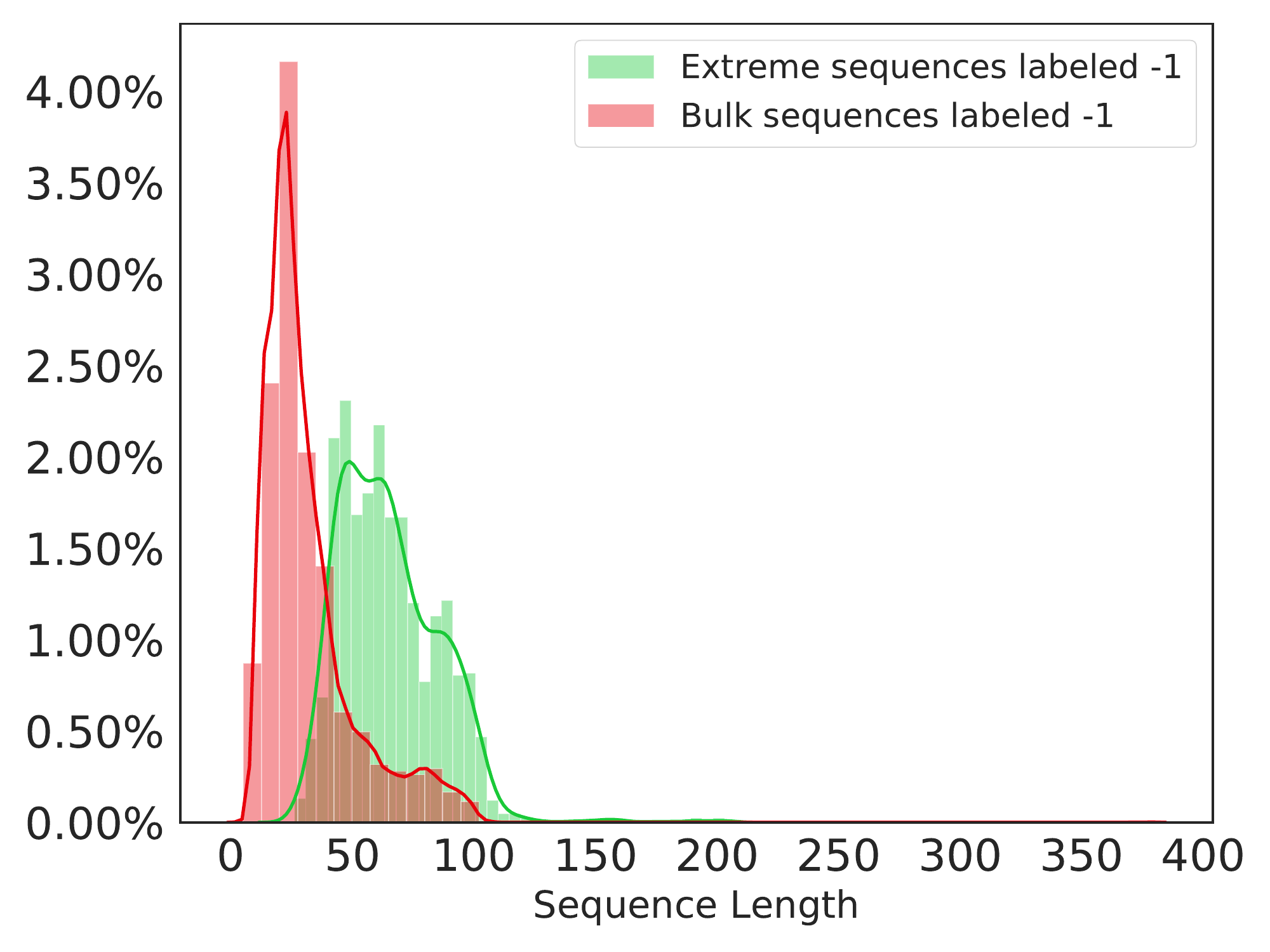}
    \caption{\textit{Amazon} - labeled $-1$}
    \label{fig:Amazon-1}
\end{subfigure}
\caption{Histograms of the samples' sequence 
length for \textit{Yelp} dataset (Figure~\ref{fig:Yelp+1} and Figure~\ref{fig:Yelp-1}) and \textit{Amazon}  (Figure~\ref{fig:Amazon+1} and Figure~\ref{fig:Amazon-1}). The number of sequences 
in the bulk is approximately $3$ times the number of extreme sequences 
for each dataset $10000$ sequences 
are considered and extreme region contains approximately $3000$ sequences 
.}
\label{fig:lengrh}
\end{figure}

Figure~\ref{fig:lengrh}  provides additional insight about the magnitude of the shift in sequence  length between extremes in the \HTalgo\ representation and non extreme samples.  Even though the histograms overlap (so that two different sequences of same length may be regarded as extreme or not depending on other factors that are not understood yet), there is a visible  shift in distribution for both \textit{Yelp} and \textit{Amazon} datasets,  both for the positive and negative class in the classification framework for  sentiment analysis. 
Kolmogorov-Smirnoff tests between the length distributions of the two considered classes for each label were performed, which allows us to reject the null hypothesis of equality between distributions, as the maximum $p$-values is less than  $0.05$.

\paragraph{Experimental conclusions} 
We summarize the empirical findings of this section: 
\begin{enumerate}
    \item An `extreme' text sequence in \HTalgo\ representation is more likely to have a greater length (number of tokens) than a non extreme one. 
    
    \item Positive correlation between the BERT norm and the \HTalgo\ norm indicates that  a large sample in the BERT representation is likely  to have a large norm in the \HTalgo\ representation as well: 
    the learnt representation \HTalgo\ taking BERT  as input keeps invariant  (in probability) the  ordering implied by the norm. 
    \item A consequence of the two above points is that long sequences tend to have a large norm in BERT. 
    
    \item  Extreme text samples (regarding the  BERT norm or the  \HTalgo\ norm) tend to be harder to model than non-extreme ones.
    
    \item Since extreme texts are harder to model and also somewhat harder to classify in view of the BERT classification scores reported in Table~\ref{tab:overall_results}, there is room for improvement in their analysis and  it is no wonder that a method dedicated to extremes  \textit{i.e.} relying on EVT  such as \HTalgo\ outperforms  the baseline.  

\end{enumerate}

\end{document}